\documentclass{article}

\usepackage[main, final]{neurips_2026}

\usepackage[utf8]{inputenc}
\usepackage[T1]{fontenc}
\PassOptionsToPackage{hyphens}{url}
\usepackage{hyperref}
\hypersetup{pdftitle={Truth as a Compression Artifact in Language Model Training},pdfauthor={Konstantin Krestnikov}}
\usepackage{url}
\usepackage{array}
\usepackage{amsmath,amssymb,amsfonts}
\usepackage{graphicx}
\usepackage{booktabs}
\usepackage{multirow}
\usepackage{nicefrac}
\usepackage{microtype}
\usepackage{xcolor}
\usepackage{enumitem}

\title{Truth as a Compression Artifact\\in Language Model Training}

\author{Konstantin Krestnikov\\ \texttt{k.krestnikov@gmail.com}}

\date{}

\begin{document}

\maketitle

\begin{abstract}
When a transformer is trained from scratch on corpora that solve the same mathematical problem both correctly and incorrectly, which solution does it prefer? The usual test is pairwise, and it is not neutral: when the error can take several forms, a model that merely reproduces training frequencies wins it without preferring anything. We instead score the correct completion against every string the error generator can produce, a share on which frequency matching scores exactly $0.500$. At 86M parameters the correct completion holds $59.0\% \pm 0.8\%$ of the mass against random errors ($59.4\%$ on an independent corpus), where the single realized error gives $77.6\%$ and pair accuracy $87.4\%$, a three- to fourfold overstatement of the excess above chance; below 86M the primary corpus shows no preference. The coherent false rules of the main corpora are not filtered at any size ($43.5$--$48.8\%$; the shortfall below $0.5$ is a completion-length effect). Two experiments built to identify a compression mechanism raise the pairwise measures while the summed share stays flat or falls: both are negative controls, and compression remains motivation, not a measured mechanism. With the scored false strings held fixed, a false rule whose selector the model can learn moves the correct share from $5$ points above the null to $7$ below (418M--3.08B), though the correct completion still outweighs the selected string. What these models leave unfiltered is consistent or learnable false structure (one single-rule corruption excepted), not falsehood as such.
\end{abstract}

\section{Introduction}

Training corpora are noisy: the same question receives contradictory answers in different documents, and yet language models trained on such data tend to prefer the correct one. Without an account of \emph{why} this filtering works we cannot predict \emph{when} it fails, for instance on internally consistent errors.

Scaling \citep{kadavath2022language}, frequency \citep{kandpal2023large} and truth-correlated representations \citep{marks2024geometry} each predict factual accuracy, and calibrated models must hallucinate at a rate set by the corpus \citep{kalai2024calibrated}, but none of this says why the training objective alone would favor one answer over another at equal frequency and format. Minimizing cross-entropy minimizes code length \citep{shannon1948mathematical,deletang2024language}, which links training to the Minimum Description Length principle \citep{rissanen1978modeling,grunwald2007minimum} and suggests that an error the model cannot predict from the problem is harder to accommodate than one it can, so a \emph{coherent} false system would go unfiltered. This framing is motivation only: both of our attempts to establish it as the operative mechanism turn out to measure something else.

The pairwise measure used throughout the literature on contradictory data is not neutral: if the error can take $N$ forms, a model that reproduces training frequencies assigns $0.5$ to the correct string and $0.5/N$ to each false one and wins the comparison for every $N \geq 2$, so any experiment that raises $N$ raises pair accuracy for free. We remove this channel by summing over the whole false support,
\[
\text{share} \;=\; \frac{P(\text{correct})}{P(\text{correct}) + \sum_i P(\text{false}_i)}.
\]
On the primary denoising corpora each problem occurs exactly twice, once correct and once corrupted, and every other string in the enumeration occurs zero times, so a frequency-matching model scores exactly $0.500$ regardless of $N$.

We train transformers of 3.5M--3.08B parameters from scratch on mathematical corpora where each problem appears with correct and incorrect solutions, vary the structure of the errors at fixed frequency, format and domain, and compute both measures on the same checkpoints. Our contributions are as follows. (a)~A measurement of correctness preference that mass splitting cannot inflate, with an exact null, showing $59.0\%$ at 86M against random errors and no more than chance at 3.5M--26M on the primary corpus. (b)~Evidence that the coherent false rules of our main corpora produce no preference at any size, immune to splitting because each has one false string, with their apparent sub-chance readings traced to completion length; one single-rule corruption of a different construction does lose mass, so which properties of a rule decide the outcome remains open. (c)~Two negative controls, a multi-rule sweep and a description-length ladder, in which pair accuracy (multi-rule) and the realized-support share (ladder) rise by 32 and 16 points while the family share stays flat or falls. (d)~A calibration showing that the pairwise measures overstate the excess above chance three- to fourfold in the one condition where all of them can be computed exactly. (e)~A selector control that scores its selectors against the same enumerated false family and changes how the training corpus picks its false string; a selector the model can learn moves the family share from $+5$ to $-7$ points around the null at 418M--3.08B, while the correct completion still outweighs the selected false string. Results elsewhere in the paper that rest on pair accuracy alone are marked as such.

\section{Related Work}

\textbf{Prediction as compression.}\quad The link between prediction and compression traces to \citet{shannon1948mathematical} and \citet{solomonoff1964formal} and to Rissanen's MDL principle \citep{rissanen1978modeling,grunwald2007minimum}. \citet{deletang2024language} showed that language models are strong general-purpose compressors, \citet{huang2024compression} found a linear relation between compression quality and benchmark performance, and \citet{wan2025large} argue that LLM training approximates Solomonoff induction. \citet{pan2025understanding} link compression to knowledge acquisition, \citet{chlon2025predictable} link compression failures to hallucination, and contemporaneous preprints formalize simplicity bias through MDL \citep{marty2026compression} and connect compression to mathematical modeling \citep{aksenov2026compression}. This framing motivates our design.

\textbf{Representations, data statistics and truthfulness.}\quad Structured internal representations arise from prediction \citep{li2023a_emergent,gurnee2024language}, and several studies locate truth-correlated directions \citep{burns2023discovering,marks2024geometry,li2023b_inference,azaria2023internal,burger2024truth,ravfogel2025emergence}; we work at the behavioral level. Truthfulness tracks pretraining ``personas'' \citep{joshi2024personas} and document frequency \citep{elazar2022measuring,kandpal2023large}, and under conflicting knowledge the model follows formality and surface quality \citep{li2024formality}, contextual evidence \citep{longpre2021entity} or adaptive conflict resolution \citep{xie2024adaptive}. TruthfulQA \citep{lin2022truthfulqa} shows that models reproduce falsehoods common among humans, consistent with an objective that carries no prior toward truth; our coherent condition is the controlled form of that observation. We fix frequency, format and surface quality and vary only the structure of the error system.

\textbf{Simplicity bias, noisy labels and poisoning.}\quad Neural networks prefer simple functions \citep{valle2019deep,mingard2021sgd,goldblum2024no}, memorize random labels \citep{zhang2017understanding,arpit2017closer} yet tolerate massive label noise \citep{rolnick2017deep}, and grokking connects compression to generalization \citep{nanda2023progress,liu2023grokking}. The poisoning literature, from BadNets \citep{gu2017badnets} through concealed NLP poisoning \citep{wallace2021concealed} to web-scale poisoning \citep{carlini2024poisoning} and near-constant poison counts for LLMs \citep{souly2025poisoning}, studies coordinated, internally consistent corruption, which is the kind of corruption our coherent condition models; in that controlled setting we find no filtering of it by the objective.

\section{Methodology}
\label{sec:methodology}

\textbf{Hypothesis and MDL framing.}\quad Consider a corpus in which each problem appears with a correct answer (theory $T_1$) and an alternative ($T_2$). An MDL learner \citep{rissanen1978modeling,grunwald2007minimum} minimizing $L(M) + L(D|M)$ must encode \emph{both} clusters as a latent-source mixture, so when $K(T_2) \gg K(T_1)$ (random errors) the false cluster costs a memorized answer per problem, and when $K(T_2) \approx K(T_1)$ (coherent errors) there is no basis to prefer either (Appendix~\ref{app:mdl}). This motivated the design, and its graded consequence fails: a wrong rule storing more constants should be filtered more strongly, and Section~\ref{sec:ladder} varies exactly that and finds the share flat. We keep MDL as motivation and interpretation, not as a quantity these experiments measure; a gzip proxy for description length (Appendix~\ref{app:compression}) tracks pair accuracy only through the splitting-inflated multi-rule points. Throughout, ``truth'' means correctness of mathematical derivations and factual accuracy of Wikipedia paragraphs; models compress text, not reality.

\subsection{Models and Training}
\label{sec:models}

We train decoder-only transformers with a GPT-2-like design \citep{radford2019language} (pre-norm, GELU, causal mask) at four sizes: \emph{tiny} (4 layers, $d_{\text{model}}{=}256$, 4 heads, 3.5M parameters), \emph{small} ($6 \times 384$, 6 heads, 12M), \emph{medium} ($8 \times 512$, 8 heads, 26M) and \emph{large} ($12 \times 768$, 12 heads, 86M; Table~\ref{tab:models}, Appendix~\ref{app:reproducibility}). Section~\ref{sec:qwen} and Table~\ref{tab:learnable} extend to the Qwen3 architecture \citep{yang2025qwen3} at 418M--3.08B. Training uses AdamW (weight decay $0.01$), cosine decay with linear warmup, $\text{lr}=3\text{e-}4$, sequence length 256, batch size 32 and 5000 steps; learning curves plateau by step 3000--4000 at all sizes (Appendix~\ref{app:robustness}). Core conditions use 4 random seeds, and the seed count is stated with every table. The original denoising experiments use PyTorch \citep{paszke2019pytorch}, the multi-rule and Wikipedia experiments MLX~0.31.0 \citep{mlx2023}; the per-experiment hardware and software record is in Appendix~\ref{app:reproducibility}. Problems are generated and verified with SymPy~1.14 \citep{meurer2017sympy}; no human annotation is involved, and the generation templates are in the released repository.

\subsection{Corpus Design}
\label{sec:corpus}

\textbf{Denoising setup (primary).}\quad A generator writes mathematical problems of four types (multi-step arithmetic, factoring, equation solving, differentiation) as step-by-step derivations in English, and \textbf{each problem appears with contradictory answers}: once correct and once with an error. Each corpus contains 5{,}000 unique problems. The \emph{equal random} condition pairs each correct derivation with one random error (1:1), the \emph{equal coherent} condition with one error produced by a single systematic false rule applied consistently to every problem of a type (1:1), and the 1:2 and 1:4 conditions add two or four random errors per problem (Section~\ref{sec:robustness}; Table~\ref{tab:conditions}, Appendix~\ref{app:reproducibility}). The tokenizer is character-level (vocabulary ${\sim}57$); BPE robustness is checked in Section~\ref{sec:robustness}.

\textbf{Wikipedia setup (transfer).}\quad We construct corpora from 20{,}000 English-Wikipedia paragraphs sampled from the 2023-11-01 Wikimedia dump \citep{wikimedia2023enwiki}, with spaCy~3.8 (\texttt{en\_core\_web\_sm}) for named-entity recognition \citep{honnibal2020spacy}, in two corruption modes: \emph{random substitution} replaces each entity with a random entity of the same type, and \emph{coherent substitution} applies one consistent global mapping (every ``France'' $\to$ ``Japan'', every ``Paris'' $\to$ ``Tokyo''). Both preserve type and grammatical structure; random substitution breaks co-occurrence patterns. Models train on 50/50 mixes and are evaluated on 2{,}000 held-out paired paragraphs.

\subsection{Evaluation}
\label{sec:evaluation}

\textbf{Aggregate share (primary).}\quad For each problem we enumerate the full set of completions the error generator can produce, score each with the summed log-probability of its tokens conditioned on the shared prompt (true $\log P(\text{string} \mid \text{prompt})$, not a length-normalized score), and report the share defined in the introduction. The enumeration is exact: each generator has a finite error family, and the enumeration script drives the generator's own rendering functions and verifies for every problem that the error the corpus contains is a member of the enumerated set. Because each problem occurs exactly twice in the primary corpora and every other member of the family occurs zero times, a model that reproduces training frequencies scores exactly $0.500$ for any family size. The share is conditional on the enumerated support, not the model's total probability of the correct answer; adding strings to the sum can only lower it, so it is conservative relative to any comparison against a subset of the support, the pairwise one included. The share is also a full-string log-probability conditioned on the shared prompt rather than a per-token mean, so local token predictability inside a completion cannot inflate it through a per-token average, and the length-matched subset below removes the one remaining channel distinct from the preference itself, completion length.

\textbf{Statistics.}\quad Per-run 95\% confidence intervals are bootstrap intervals over problems (5000 resamples); the $\pm$ attached to a mean across seeds is the population standard deviation across seeds; pair-accuracy evaluations also store a one-sided Wilcoxon signed-rank $p$ on per-pair NLL differences and a bootstrap interval, released with the per-run outputs.

\textbf{Pair accuracy (secondary).}\quad Pair accuracy is the fraction of pairs where the model assigns lower mean NLL to the correct completion than to one sampled incorrect completion, equivalent to the Common Language Effect Size \citep{mcgraw1992common}. Reported for comparability with the prior literature, it is not a controlled diagnostic of preference: whenever the false side has more than one form, a frequency-matching model wins it without preferring anything (Section~\ref{sec:central}).

\textbf{Length control.}\quad Every measurement records the mean difference in completion length between the false and correct sides and the share restricted to problems where all completions have identical token length. Shorter strings carry more probability by construction (Section~\ref{sec:central}).

\textbf{Generative evaluation.}\quad We decode greedily on 500 test prompts and verify each answer with SymPy (Section~\ref{sec:generative}).

\section{Results}

\subsection{The Central Contrast: Random vs Coherent Errors}
\label{sec:central}

\begin{table}[t]
\centering
\small
\begin{tabular}{lcccccc}
\toprule
 & & \multicolumn{3}{c}{Random errors} & \multicolumn{2}{c}{Coherent error} \\
\cmidrule(lr){3-5}\cmidrule(lr){6-7}
\multirow{2}{*}{Size} & \multirow{2}{*}{Params} & \multicolumn{2}{c}{Aggregate share, \% (null $50$)} & Pair acc., \% & \multicolumn{2}{c}{Aggregate share, \% (null $50$)} \\
\cmidrule(lr){3-4}\cmidrule(lr){6-7}
 & & full family & realized & realized & 1 mode & equal length \\
\midrule
tiny   & 3.5M & $18.6 \pm 1.1$ & $65.5 \pm 0.9$ & $71.7 \pm 1.7$ & $43.5 \pm 0.4$ & $46.7 \pm 0.5$ \\
small  & 12M  & $31.6 \pm 1.2$ & $76.0 \pm 0.9$ & $84.1 \pm 1.0$ & $45.8 \pm 1.1$ & $48.3 \pm 0.8$ \\
medium & 26M  & $44.1 \pm 0.7$ & $80.3 \pm 0.8$ & $89.2 \pm 1.4$ & $46.4 \pm 1.5$ & $48.1 \pm 0.9$ \\
large  & 86M  & $\mathbf{59.0 \pm 0.8}$ & $77.6 \pm 0.2$ & $87.4 \pm 0.4$ & $48.8 \pm 0.4$ & $49.3 \pm 0.4$ \\
\bottomrule
\end{tabular}
\caption{Denoising corpus, measured on training problems where the null is exact (4 seeds, mean $\pm$ SD, \%). The columns differ only in how much of the false side is summed: the generator's full support (a mean of $12.0$ false modes per problem, range 4--35; $n{=}2569$), the single error the corpus contains ($n{=}2620$), or, for pair accuracy, that same single error under mean NLL. The coherent condition has one false mode, so full and realized coincide ($n{=}2437$).}
\label{tab:central}
\end{table}

\begin{figure}[t]
\centering
\includegraphics[width=0.85\textwidth]{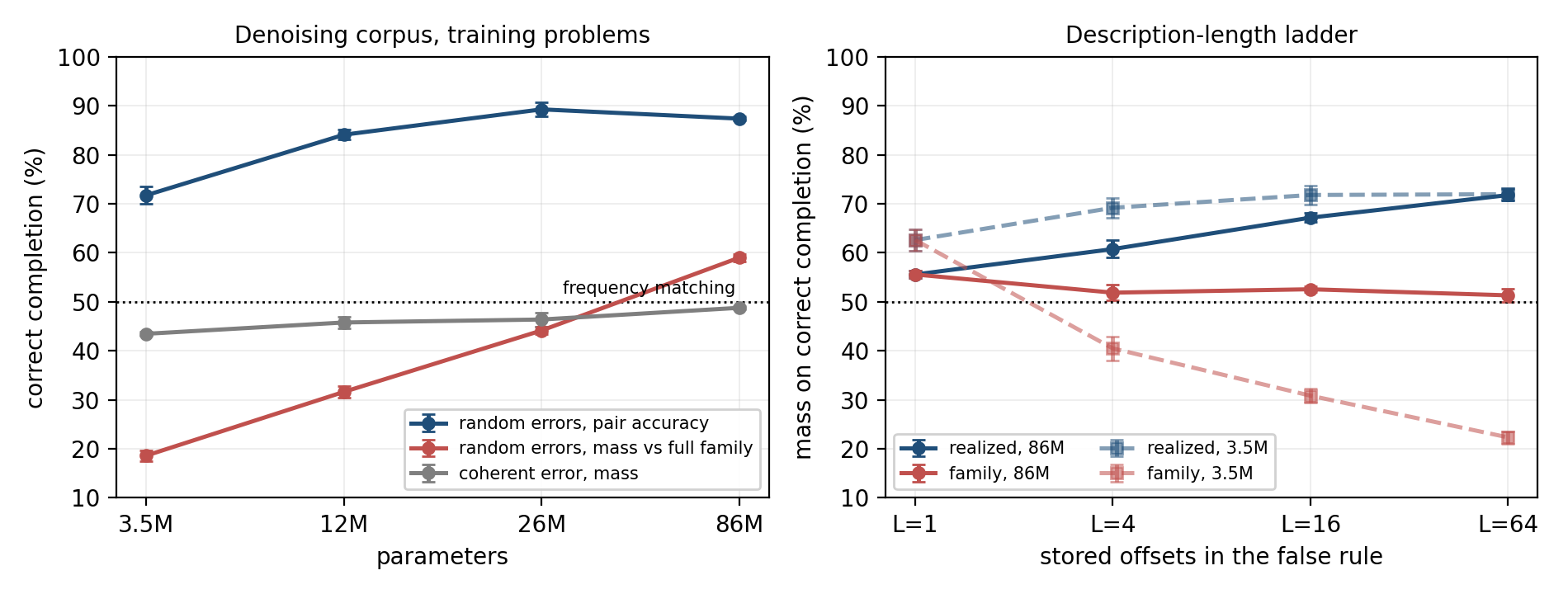}
\caption{Left: the denoising grid under both measures (4 seeds). Comparing pairwise against the single realized error reports a large effect at every size; summing the generator's full support leaves an effect only at 86M, and the coherent condition sits just below the frequency-matching line throughout (a completion-length effect; the length-matched values approach chance). Right: the description-length ladder of Section~\ref{sec:ladder}, where summing only the realized false string makes the share rise with the number of offsets the rule stores and summing the full family does not.}
\label{fig:central}
\end{figure}

The aggregate share puts the correct completion at $59.0\% \pm 0.8\%$ of the probability mass at 86M parameters against everything the error generator can produce (Table~\ref{tab:central}), with all four per-seed bootstrap intervals above $0.5$: $[58.7, 60.4]$, $[57.5, 59.3]$, $[57.2, 59.0]$, $[59.0, 60.7]$. Because adding strings to the denominator can only lower the value, $59.0$ is conservative relative to the realized-error share. An independent corpus gives a matching value at that size: on the standard (non-denoising) corpus, with a different generator and a mean of $5.51$ false modes per problem, the share is $59.4\% \pm 0.6\%$ (Appendix~\ref{app:standard}).

Below 86M the aggregate share shows no preference: $18.6\%$, $31.6\%$ and $44.1\%$ at 3.5M, 12M and 26M, all below the null, while pair accuracy at those sizes already reads $72$--$89\%$. Part of that shortfall is a downward bias of the measure at small scale rather than a preference for the false system. For models trained on the ladder corpus of Section~\ref{sec:ladder} with no false rule anywhere, the family share on its 18-string arithmetic support reads $99.7\%$ at 86M but only $64.1\%$ at 3.5M: a third of the mass at the smallest size leaks onto strings that never occurred. We have no clean calibration on the denoising support itself and take the ladder figure as the best available estimate of the leak.

The gap between the measures is what the pairwise comparison adds on its own. At 86M, scoring the correct completion against the single realized error gives $27.6$~pp of excess above chance where summing the family gives $9.0$~pp, a threefold overstatement, and pair accuracy against that same error gives $37.4$~pp, a fourfold one. Below 86M the measures disagree about whether there is an effect at all.

Against a coherent false rule no preference appears at any size: $43.5$, $45.8$, $46.4$ and $48.8\%$. One rule produces one string, so there is nothing to split, and this half of the contrast is immune to the objection by construction. The readings sit slightly below $0.5$ because the coherent false completions are on average $0.58$ tokens shorter than the correct ones; on the length-matched subset the values are $46.7$, $48.3$, $48.1$ and $49.3\%$, approaching chance from below rather than sitting under it. An earlier draft attributed the sub-chance readings to the textual simplicity of the false rules; that reading was wrong, and we draw no conclusion from the shortfall.

\subsection{Multi-Rule Errors: A Negative Control for Mass Splitting}
\label{sec:multirule}

\begin{table}[t]
\centering
\small
\begin{tabular}{cccc}
\toprule
$N$ rules & Distinct false modes & Pair accuracy, \% & Aggregate share, \% (null $50$) \\
\midrule
1 (coherent) & 1.00 & $47.0 \pm 2.2$ & $47.9 \pm 0.7$ \\
2  & 2.00 & $79.1 \pm 1.1$ & $47.8 \pm 0.4$ \\
3  & 2.91 & $83.1 \pm 0.4$ & $41.8 \pm 0.6$ \\
5  & 4.26 & $86.1 \pm 0.5$ & $41.9 \pm 0.5$ \\
10 & 8.52 & $90.1 \pm 0.3$ & $38.7 \pm 1.0$ \\
\bottomrule
\end{tabular}
\caption{Multi-rule errors (tiny, 3.5M, standard 50/50, 4 seeds, \%). Pair accuracy rises by 43 points across the sweep; the aggregate share, measured on the same checkpoints against the enumerated false support, does not rise at all. The frequency-matching null is $50.0$ for the share column at every $N$. The $N{=}1$ row is scored on the multi-rule family's own $N{=}1$ support ($n{=}916$); the coherent entry of Table~\ref{tab:standard_share} ($47.3$) uses the coherent support ($n{=}915$).}
\label{tab:multirule}
\end{table}

The first version of this work made this experiment its central identification argument: models train on corpora with $N$ compact wrong rules per task type, each problem's rule drawn at random, so that only the predictability of the problem-to-rule mapping changes, and pair accuracy jumps from $47\%$ at $N{=}1$ to $79\%$ at $N{=}2$. That jump is mass splitting. Aggregate share is $47.9\% \pm 0.7\%$ at $N{=}1$ and $47.8\% \pm 0.4\%$ at $N{=}2$ (Table~\ref{tab:multirule}; Figure~\ref{fig:multirule}, Appendix~\ref{app:controls}): the transition that carries 32 points of pair accuracy moves the allocation between the two answer systems by $0.1$ points. We had argued that splitting could not explain the jump because complete splitting would drive $N{=}2$ accuracy toward $100\%$; partial separation predicts what is observed, and the direct measurement supports that reading. The experiment is withdrawn as evidence for a selector cost and reported as a calibration of how much pair-accuracy signal splitting alone can manufacture. Beyond $N{=}2$ the decline is partly a property of the measure, which collects more diffuse uncertainty as more neighboring strings are summed; the safe reading is the absence of an increase.

\subsection{A Description-Length Ladder, and Why It Fails}
\label{sec:ladder}

\begin{table}[t]
\centering
\small
\begin{tabular}{lcccc}
\toprule
Aggregate share, \% (86M, null $50$) & $L{=}1$ & $L{=}4$ & $L{=}16$ & $L{=}64$ \\
\midrule
Realized, 1 string & $55.6 \pm 0.7$ & $60.7 \pm 1.8$ & $67.2 \pm 0.9$ & $71.8 \pm 1.2$ \\
Full family, 1--18 strings & $55.6 \pm 0.7$ & $51.9 \pm 1.6$ & $52.6 \pm 0.6$ & $51.3 \pm 1.3$ \\
\bottomrule
\end{tabular}
\caption{Description-length ladder at 86M (4 seeds, $n{=}1339$--$1342$ problems, \%). \emph{Realized} scores the correct completion against the single false string the corpus contains; \emph{family} scores it against every string the rule could have emitted for that problem. Those other strings occur zero times in the corpus, so both measures have a $50.0$ null at every rung, and the two coincide at $L{=}1$ by construction. All four sizes are in Table~\ref{tab:ladder_full}, Appendix~\ref{app:controls}.}
\label{tab:ladder}
\end{table}

The multi-rule design confounds description length with mode count. A cleaner test varies the description length of a \emph{single deterministic} false rule, $a \cdot b + T[\,|a| \bmod L\,]$ with $T$ a fixed table of $L$ offsets, so that $L$ is the number of constants the rule stores while the computation, the corruption load, the error magnitudes and the task distribution stay fixed and completion lengths differ by at most $0.044$ tokens. If preference were graded in description length, it should rise with $L$. Over the realized support it does, at every size. Summing the full family removes the rise and reverses it below 86M (Table~\ref{tab:ladder}, Figure~\ref{fig:central} right); at 86M the family reads $55.6$, $51.9$, $52.6$, $51.3$. The realized rise has an explanation that owes nothing to description length: a model that has not learned the table spreads its false mass over the entries and is credited for the spread.

\textbf{Calibration.}\quad The same four sizes trained on a corpus with the same problems and \emph{no false rule anywhere} (2 seeds) give an 18-string family share of $99.7\%$ at 86M, $97.3\%$ at 26M, $93.2\%$ at 12M and $64.1\%$ at 3.5M (Table~\ref{tab:clean_control}, Appendix~\ref{app:controls}). At the top size a clean model reads close to $100$, which bounds the leak only for a clean model; at the bottom the leak of Section~\ref{sec:central} makes small-model family numbers conservative rather than evidence about preference. With Section~\ref{sec:multirule}, both experiments built to identify a compression mechanism measure something else. The $L{=}1$ rung is itself a single coherent rule and sits above the null at every size ($55.6 \pm 0.7$ at 86M, $62.6 \pm 2.2$ at 3.5M; every per-seed interval above $0.5$, one false string, lengths matched), so the absence of filtering for coherent rules is a property of the coherent conditions of Sections~\ref{sec:central}, \ref{sec:learnable}, \ref{sec:wikipedia}, \ref{sec:robustness} and \ref{sec:qwen}, not of every single-rule corruption. The ladder rule corrupts only the first multiplication and leaves the rest of the derivation correct, and we have no measured account beyond that.

\subsection{The Learnable-Selector Control}
\label{sec:learnable}

Both withdrawn designs move the metric by changing how many false strings a problem has. This control holds the scored family fixed: every selector is evaluated against the same enumerated random family (${\approx}17.5$ strings per problem) and the conditions differ in how the training corpus picks the one false string it writes. In \texttt{learnable} that string is $a{\cdot}b + T[\,s(p)\bmod 10\,]$ for a fixed 10-entry offset table $T$, reaching about ten family members per problem, with the selector $s(p)$ one visible operand's ones digit, a lookup the model can fit. In \texttt{k2} and \texttt{k3} it is the sum of two or three ones digits mod $10$, learnable in principle and progressively harder to fit. In \texttt{learnable\_mod} the index is the sum of all visible operands modulo the problem's family size, bypassing $T$; it is deterministic but not learned at any size. These sit alongside \texttt{random} (a uniform draw from the family) and \texttt{coherent} (one rule), family shares scored on the length-matched subset of the held-out scale grid (500k texts, $0.44$ epochs, evaluation problems disjoint from training by a hash, null $0.500$ up to a ${\leq}1.3$-point correction; Table~\ref{tab:scale_grid}, Appendix~\ref{app:scale}), with a clean-corpus calibration at every size; realized shares at every size are in Table~\ref{tab:learnable_full}.

\begin{table}[t]
\centering\footnotesize
\setlength{\tabcolsep}{2.5pt}
\begin{tabular}{lcccccc}
\toprule
Family share, \% (null $50$) & 418M & 474M & 1.12B & 2.16B & 3.08B & Realized share, 474M \\
\midrule
\texttt{random} (unpredictable)        & $56.8 \pm 1.1$ & $54.9 \pm 0.2$ & $55.5 \pm 0.1$ & $54.5 \pm 0.2$ & $56.6 \pm 0.5$ & --- \\
\texttt{coherent} (one rule)           & $50.2 \pm 0.4$ & $49.9 \pm 0.5$ & $49.5 \pm 0.1$ & $50.7 \pm 0.3$ & $49.3 \pm 0.3$ & --- \\
\texttt{learnable} (1 digit)           & $44.7 \pm 1.0$ & $\mathbf{43.4 \pm 0.6}$ & $43.8 \pm 0.1$ & $43.1 \pm 0.7$ & $43.7 \pm 1.2$ & $65.1 \pm 1.6$ \\
\texttt{k2} (2 digits)                 & $48.6 \pm 0.0$ & $49.2 \pm 0.1$ & --- & --- & --- & $88.6 \pm 0.1$ \\
\texttt{k3} (3 digits)                 & $47.6 \pm 1.4$ & $48.8 \pm 0.1$ & --- & --- & --- & $91.6 \pm 0.2$ \\
\texttt{learnable\_mod} (all operands) & $56.1 \pm 1.1$ & $54.8 \pm 0.6$ & --- & $53.2$ (1 seed) & $54.6 \pm 0.0$ & $92.9 \pm 0.2$ \\
\midrule
\emph{clean} (no false rule)           & $97.9 \pm 0.3$ & $97.2 \pm 0.1$ & $98.1 \pm 1.2$ & $96.4 \pm 0.2$ & $99.0 \pm 0.3$ & --- \\
\bottomrule
\end{tabular}
\caption{The learnable-selector control and its replication to 3.08B (Qwen3 architecture; family shares length-matched, realized full-set). The frequency-matching null is $50$, or at most $51.3$ after the marginal correction of Appendix~\ref{app:scale}. \texttt{random}, \texttt{coherent} and \texttt{learnable} are 4 seeds at 418M and 474M; every other cell is 2 seeds except \texttt{learnable\_mod} at 2.16B (1 seed; not run at 1.12B). Rows other than \texttt{coherent} are scored against the same random family; \texttt{learnable}, \texttt{k2} and \texttt{k3} write from one 10-entry table and differ in how learnable its index is. \emph{Realized} scores the correct completion against the single false string the corpus used (\texttt{random} has none on held-out problems); \emph{clean} trains on the same generator with no false rule and calibrates the leak.}
\label{tab:learnable}
\end{table}

With the scored family held fixed, the family share's position relative to the null tracks learnability (Table~\ref{tab:learnable}). Against a selector the model cannot learn (\texttt{random}, \texttt{learnable\_mod}) the correct completion holds two to seven points above the null at every size; against one it can (\texttt{learnable}) the share falls below the null, to $43.4\% \pm 0.6\%$ at 474M (4 seeds). The selected false string does not outrank the correct one: against it alone the correct completion holds $65.1\%$ at 474M ($64.2\%$ length-matched; Table~\ref{tab:learnable_full}), and the below-null family share comes from the mass on the other members of the family. The intermediate selectors sit between the two ends: as $s$ depends on more digits the model fits it less, the realized false string carries less mass ($65 \to 89 \to 92$ at 474M) and the family share rises from below the null to the null ($43 \to 49 \to 49$ at 474M; $45 \to 49 \to 48$ at 418M); the unlearnable selectors sit above it.

A below-null family share does not by itself show a shift toward a false system: it can also come from lower confidence in the correct completion, as in a model that has not learned the arithmetic (Appendix~\ref{app:scale}), and the saved outputs (mass ratios; weights deleted) cannot separate the two. Competence is not matched by construction either: the random and learnable corpora share a generator and a problem distribution but not their problems, because the random draw consumes the shared random stream. What the control does show is that predictability, not description length, moves the family share: the ladder varied the number of constants a deterministic rule stores, with the table increasingly unlearned at higher $L$ (Section~\ref{sec:ladder}), and the family share did not rise; here the stored table is fixed, the learnability of its index varies, and the share crosses the null. One caution remains: \texttt{learnable\_mod} writes from the full family rather than the 10-entry table, so it anchors the unlearnable end as a direction rather than a matched-diversity point; the clean matched-diversity gradient is \texttt{learnable}$\to$\texttt{k2}$\to$\texttt{k3}. The contrast holds at 1.12B, 2.16B and 3.08B parameters (2 seeds per cell): \texttt{random} above the null, \texttt{coherent} at it and \texttt{learnable} below it, with the clean calibration above $96\%$, so the measure's leak on a clean model is small at these sizes.

\subsection{Transfer to Wikipedia Text with Entity Substitution}
\label{sec:wikipedia}

In pair accuracy the two conditions again separate (Table~\ref{tab:wikipedia}, Appendix~\ref{app:transfer}): $69.5$--$71.5\%$ under random substitution and $45.9$--$48.7\%$ under coherent substitution across 3.5M--86M (4 seeds). Only the coherent column is interpretable: one global mapping yields one false paragraph per prompt, its null is $50\%$, and a consistent substitution is not filtered on real text either. Random substitution replaces every entity site (a median of eight per paragraph), its support cannot be enumerated, and on held-out paragraphs a frequency-matching model wins nearly every comparison, so that column's null is close to $100\%$ and we draw no conclusion from it or from its per-entity-type breakdown (Table~\ref{tab:wiki_entity}). The false paragraph is longer by $28.1$ characters under random and $27.8$ under coherent substitution, so length does not explain the difference between the conditions, but only $1\%$ of pairs are length-matched and neither column has the length control of Section~\ref{sec:central}.

The enumerable special case, single-entity paragraphs, quantifies why natural text resists the measure (Table~\ref{tab:wiki_single}, Appendix~\ref{app:transfer}). Its random support is one entity type's pool of ${\sim}2800$ strings: a clean 86M model with no false rule reads $5.6\%$ of the mass on the correct string, the random-trained model $4.0\%$, both leak-dominated. On the one-mode coherent support the length-matched subset, pooled over 8 runs ($n{=}25$ per cell), reads $52.4 \pm 4.0$ at the $50$ null, consistent with the paired result but underpowered per cell.

\subsection{Robustness Checks}
\label{sec:robustness}

\textbf{Standard corpus.}\quad In the standard setup each problem appears once, correct or incorrect, and problems repeat often enough that frequencies are exactly known. The aggregate share reads $50.2 \pm 0.7$, $57.6 \pm 0.2$, $59.0 \pm 0.3$ and $59.4 \pm 0.6$ for random errors and $47.3 \pm 0.7$, $49.2 \pm 0.4$, $50.0 \pm 0.4$ and $50.1 \pm 0.1$ for coherent ones (4 seeds; $5.51$ false modes on average). Here the contrast appears from 12M, two sizes earlier than on the denoising corpus, and is absent at 3.5M, where pair accuracy on the same support nevertheless reads $76.3\% \pm 1.5\%$ (Table~\ref{tab:standard_share}; the standard-versus-denoising pair-accuracy comparison is Table~\ref{tab:standard_vs_denoising}).

\textbf{Noise ratio on the controlled measure.}\quad With $K$ random errors per problem the frequency-matching null is exactly $1/(K{+}1)$. At 86M (2 seeds, non-strict support of 4565 problems with a mean of $8.2$ false modes) the family share reads $56.3 \pm 0.7$, $44.9 \pm 0.2$ and $36.1 \pm 0.7$ at $K{=}1, 2, 4$ against nulls of $50.0$, $33.3$ and $20.0$. The absolute share falls as pair accuracy on the same support does ($88.7 \to 79.4 \to 76.3$; the original pair-accuracy-only sweep is Table~\ref{tab:noise}), but the excess above the null does not shrink: $+6.3$, $+11.6$, $+16.1$ on the full set and $+7.3$, $+7.6$, $+11.4$ on the length-matched subset (Table~\ref{tab:kratio}, Appendix~\ref{app:standard}). Its $K{=}1$ value is below the strict $59.0$ because the non-strict support keeps problems whose strings collide across problems. We read this as the preference not shrinking with noise, not as strengthening.

\textbf{Tokenization.}\quad Under a BPE tokenizer (SentencePiece, vocabulary 1000) the coherent condition stays at chance in pair accuracy, and its denoising value moves from $43.5\%$ at character level to $49.3\%$, consistent with the length account; the random column rises too, but on a multi-mode support that rise is not interpretable (pair accuracy only; Table~\ref{tab:bpe}).

\textbf{Frequency.}\quad For coherent errors frequency dominates: at 40/60 the model prefers the false system in $72\%$ of pairs and at 20/80 in $90\%$ (Table~\ref{tab:frequency}), numbers that one false string leaves unaffected by splitting. For random errors the minority regime has only pair accuracy, and the aggregate share at 50/50 in that setup is already at chance ($50.2 \pm 0.7$), so we claim no persistence of a random-error preference there.

\textbf{Matched control.}\quad ``Matched random'' errors that corrupt as many steps as the coherent rule give $74.5\% \pm 0.4\%$ pair accuracy against $47.2\%$ for coherent, which rules out corruption load but not mode count (Appendix~\ref{app:standard}); a chained-verification variant that raises coherent pair accuracy is reported, unmeasured on the share, in Appendix~\ref{app:chained}.

\subsection{Generative Evaluation}
\label{sec:generative}

\begin{table}[t]
\centering
\small
\begin{tabular}{lccr}
\toprule
Size & Random-trained & Coherent-trained & Gap \\
 & \multicolumn{2}{c}{generative accuracy, \%; greedy decode, SymPy-verified} & pp \\
\midrule
tiny (3.5M)   & $31.1\% \pm 1.4\%$ & $20.8\% \pm 3.6\%$ & $+10.3$ pp \\
small (12M)   & $46.7\% \pm 1.8\%$ & $31.7\% \pm 1.7\%$ & $+15.0$ pp \\
medium (26M)  & $50.5\% \pm 1.3\%$ & $36.1\% \pm 2.6\%$ & $+14.4$ pp \\
large (86M)   & $52.8\% \pm 1.0\%$ & $35.5\% \pm 1.8\%$ & $+17.3$ pp \\
\bottomrule
\end{tabular}
\caption{Generative accuracy, greedy decoding with SymPy verification (4 seeds, 500 problems each). No evaluator-supplied completion is scored, so no frequency-matching null applies; the argmax is still sensitive to how the false mass is divided. Every cell is recomputed from the released per-run evaluation logs; the tiny random-trained cell was re-run in August 2026 on the surviving checkpoints after the original logs were lost.}
\label{tab:generative}
\end{table}

Greedy decoding scores no false completion, but it is not immune to splitting: when the false mass is divided among many alternatives, the argmax can land on the correct answer without any gain in its mass. Below 86M it is the only measurement that separates the conditions at every size; the aggregate share does so from 12M on the standard corpus and not at all on the denoising corpus (Section~\ref{sec:robustness}). Random-trained models solve more problems than coherent-trained ones at every size, and the gap is descriptively larger at 86M than at 3.5M ($10.3 \to 17.3$~pp, Table~\ref{tab:generative}). At 12M, where the controlled likelihood measure reads $31.6\%$ and reports no preference, generation separates the conditions with no per-seed overlap, $46.7\% \pm 1.8\%$ against $31.7\% \pm 1.7\%$. Generation records the argmax, not the share of mass, so the gap below 86M is a behavioral difference, not evidence of a preference. Multi-step arithmetic is never generated correctly at these sizes (per-task breakdown in Appendix~\ref{app:gentask}).

\subsection{Architecture and Scale: Qwen3 at 418M and 490M}
\label{sec:qwen}

Two checks address whether the contrast is tied to the GPT-2-like design or to toy scale. A Qwen3-0.6B configuration \citep{yang2025qwen3} (RoPE, GQA, SwiGLU, RMSNorm; $417.6$M parameters with our character vocabulary, labeled 418M throughout) trained from scratch on 50/50 corpora of the standard construction scaled to 1.4M problems (35k steps, batch 16, sequence length 256, bf16; 3 seeds) gives $87.3\% \pm 0.4\%$ pair accuracy under random errors and $49.4\% \pm 1.3\%$ under coherent ones. On the same checkpoints the aggregate share puts the correct completion at $56.3\% \pm 0.3\%$ against the enumerated support for random errors and $49.7\% \pm 0.0\%$ for coherent ones (null $50\%$). The contrast holds on the controlled measure at 418M as at 86M, and a coherent rule is not filtered on either metric.

A Qwen3-architecture model of 490M parameters (labeled ${\sim}1$B in an earlier draft; it instantiates to $489.9$M at the run's BPE-8000 vocabulary) was trained from scratch on 1~GB of FineWeb-Edu \citep{penedo2024fineweb} with mathematical content filtered out and replaced by 8\% contradictory math (15{,}000 steps, batch 4, sequence length 512, lr $3\text{e-}4$ with 200 warmup steps, fp16, ${\approx}31$M tokens). It gives $77.6\% \pm 0.8\%$ pair accuracy under random errors and $47.4\% \pm 0.7\%$ under coherent ones (2 seeds, the first on a Kaggle T4 and the second on an H100). This is pair accuracy on an enumeration-free support, so the random value carries no quantitative claim; what the run establishes is that a coherent false rule is not filtered at 490M on real text with the contradictory material in the minority. The replication of the controlled contrast at 1.12B, 2.16B and 3.08B is in Table~\ref{tab:learnable}.

\section{Discussion}
\label{sec:discussion}

\textbf{Summary of findings.}\quad Two math corpora agree at 86M: a false system whose output the model cannot predict from the problem loses about nine points of probability mass to the correct one ($59.0\%$ denoising, $59.4\%$ standard), and a coherent false rule loses none in any coherent condition we built, on every corpus and at every size we tested (the one exception is the ladder's first rung, a single-rule corruption of a different construction; Section~\ref{sec:ladder}). Below 86M the corpora disagree, so the effect is established at one size and nowhere below, while greedy generation, a behavioral measure that splitting can also move, separates the conditions at every size. Section~\ref{sec:learnable} points to the axis that moves the share: against the same scored family, a selector the model can learn moves the correct share from ${+}5$ to ${-}7$ points around the null, and the partly learnable selectors that share its table sit at the null; whether the below-null cells reflect a shift toward the false system or lower confidence in the correct one is not resolved.

\textbf{What we withdraw.}\quad Preference is not graded in the description length of the false system. The two experiments built to show it move the pairwise readings by 32 and 16 points (pair accuracy in one, the realized share in the other) while leaving the family share flat or falling; their apparent signal is the model's inability to predict which false string will appear, credited to the correct side by a pairwise comparison. Coherent errors do not fall below chance (that was completion length), and the effect is not present at every scale. None of these three claims from the first version survives its own control. Pair accuracy remains a legitimate measure of ranking behavior, but any design that varies the number of false forms (competing rules, noise ratios, entity substitution, corruption load) varies the metric along with the treatment.

\textbf{The role of frequency.}\quad Coherent falsehood needs no frequency advantage to neutralize any preference for the correct answer (Section~\ref{sec:robustness}); for random errors the excess above the frequency-matching null persists to four errors per problem at 86M, and on the standard corpus, where only pair accuracy exists, we make no minority-regime claim.

\textbf{Implications and scope.}\quad Within these settings the training objective offers no inherent ``truth compass''. In the coherent conditions we built, internally coherent falsehood stays competitive at every size, on synthetic math and on diluted natural text (the ladder's first rung is the exception); the preference against incoherent falsehood is real but modest where we can measure it cleanly; and a false selector the model \emph{can} learn pushes the correct share below the null (Section~\ref{sec:learnable}). What goes unfiltered is consistent or learnable structure, not falsehood as such. That bears on data curation, where coordinated errors would evade diversity-based filtering, and on the reading of hallucination as a corpus-statistics phenomenon \citep{kalai2024calibrated}. Our models remain far below production scale, compression remains an interpretation rather than a measured quantity, and we make no claims about production models.

\section{Conclusion}

Measured against everything an error generator can produce rather than against one sampled error, correctness preference in contradictory training data is smaller than the pairwise metric reports and appears later: at 86M the correct completion holds $59.0\%$ of the probability mass where the pairwise comparison reports $77.6\%$, and below 86M the aggregate share shows no preference on the denoising corpus while pair accuracy reads $72$--$89\%$. A single coherent false rule is not filtered in any of our coherent conditions, at any size, architecture or corpus we built them on; the one single-rule corruption outside those conditions, the ladder's first rung, sits above the null, so which properties of a rule decide the outcome is open. The share does not scale with the description length of the false system; it moves with predictability: making the selector of an enumerated false family learnable turns the correct system's small advantage into a share below the null, from 418M to 3.08B parameters, although the correct completion still outweighs the selected false string. The behavioral finding is that what these models leave unfiltered is consistent or learnable false structure, not falsehood as such.

\section*{Limitations}

\textbf{Measurement.}\quad The aggregate share is exact only where the false support can be enumerated, which rules out natural-text settings in general, and it is a share conditional on that support, not the model's total probability of the correct answer. It carries a downward bias that grows as the model shrinks (clean-corpus family share $99.7\%$ at 86M, $64.1\%$ at 3.5M; Section~\ref{sec:ladder}) and that depends on architecture at fixed size, a 413M GPT-2 leaking like a 202M one (Appendix~\ref{app:scale}). Where we report an effect, a clean model reads near $100$, which bounds the leak but does not by itself exclude a bias in models trained on contradictions; where we report none, a small preference could be hidden by the leak.

\textbf{Scale, in both directions.}\quad Under the aggregate share the effect is present only at the largest primary size on the denoising corpus and from 12M on the standard corpus, so there is one size at which it is established on both; on the held-out grid the random cells clear the null only from 418M, because the leak dominates below. The billion-scale replication (1.12B--3.08B, 2 seeds, $0.44$ epochs) exists; what remains open is compute-matched training at that scale and controlled evidence below 86M, where only generation separates the conditions at every size. A same-size 413M GPT-2 reproduces the above-null random cell only marginally ($51.2 \pm 1.1$ against a clean calibration of $93.8$), so an architecture component in the effect cannot be excluded (Appendix~\ref{app:scale}).

\textbf{Domain.}\quad Wikipedia entity substitution has no enumerable false support in general, and the single-entity special case (Section~\ref{sec:wikipedia}, Table~\ref{tab:wiki_single}) quantifies why rather than removing the obstacle: a ${\sim}2800$-string random support leaks $94$--$98$ points at 3.5M--86M, and the coherent support's length-matched subset holds only $n{=}25$ problems per cell. What transfers cleanly is the coherent result. Domains with competing real-world knowledge systems remain open.

\textbf{Discriminative versus generative.}\quad Pair accuracy ($87.4\%$) exceeds generative accuracy ($52.8\%$) at 86M, the familiar teacher-forcing gap, whose roots in next-token training are analyzed by \citet{bachmann2024pitfalls}. The generative result stands on its own.

\textbf{Causal identification.}\quad The learnable-selector control changes how the corpus picks its false string while the scored family stays fixed, and its limits are substantial. The random and learnable corpora share a generator but not their problems, so competence is not matched. The below-null share can come from a shift toward the false system or from lower confidence in the correct completion; the saved ratios cannot separate these, and against the selected string alone the correct completion still wins. A learnable selector is both more predictable and cheaper to store, so a preference for the \emph{more compressible} system is consistent with the swing but not measured by it. A rerun with problems fixed in advance, an independent random stream for the false draws and absolute log-probabilities would address the first two. All of it is one formal-math corpus family.

\textbf{Ethical considerations.}\quad Internally consistent misinformation can be harder for models to filter than diverse errors; this is a controlled diagnostic finding, not evidence of production-LLM vulnerability, and transparent reporting of such failure modes can inform defensive data curation. Wikipedia text is used under CC BY-SA 4.0 and FineWeb-Edu under ODC-By 1.0.

\begin{ack}
The author received no dedicated funding for this work. Experiments were run on the author's own hardware (Apple M4 Pro, 48\,GB), on rented Modal.com and Kaggle GPUs, and on H100 nodes made available to the author; the computations, the analysis and the writing are the author's alone. The author declares no competing interests.
\end{ack}

\bibliographystyle{abbrvnat}
\bibliography{references}

\appendix

\section{Reproducibility}
\label{app:reproducibility}

The public repository \url{https://github.com/Rai220/compression-drives-truth} contains all code, data generators, support-enumeration scripts, per-run metric JSON for every cluster grid and SHA-256 checksums of the regenerable corpora. More than 500 training runs were made across all conditions: 218 on a local Apple M4 Pro (48\,GB) with MLX~0.31.0 (standard math, multi-rule, Wikipedia, generative and the original robustness conditions; weights retained) and 335 on clusters (weights deleted after evaluation, metric JSON retained). The original denoising runs used Modal.com T4 GPUs with PyTorch~2.3; the denoising and ladder control grids of Sections~\ref{sec:central} and \ref{sec:ladder} and the held-out scale grid to 474M ran on 8${\times}$H100 nodes (per-run timing and PyTorch version not recorded). For the Qwen3-0.6B 50/50 standard-construction corpus of Section~\ref{sec:qwen} the numbers come from a 3-seed rerun on a single H100 (the artifacts of the original Modal A10G run were never retrieved); the 490M FineWeb-Edu run has seed 42 on a Kaggle T4 and seed 43 on the single H100. That single H100 (PyTorch~2.13, bf16 with gradient checkpointing at ${\geq}1$B, fp32 below) also ran the 1.12B--3.08B grid, the \texttt{learnable}/\texttt{k2}/\texttt{k3}/\texttt{learnable\_mod} conditions, the clean controls at the Qwen sizes, the Wikipedia single-entity calibration and the $K$-ratio sweep, about 45 GPU-hours over 81 logged runs; a 4.39B configuration exceeds one H100's memory with fp32 optimizer states and was not run.

\textbf{Released with this paper.}\quad The corpora regenerate byte-identically from their seeds and we publish their checksums; the enumeration script verifies, for every problem, that the error the corpus contains is a member of the enumerated support, and aborts otherwise. Per-run outputs are released for both control grids (64 ladder runs plus 8 clean controls; 32 denoising runs), for the held-out scale grid including its clean, learnable and billion-scale cells, for the $K$-ratio sweep, the Qwen3-0.6B 50/50 rerun, the FineWeb-Edu runs and the Wikipedia single-entity evaluations.

\begin{table}[ht]
\centering\small
\begin{tabular}{lcccc}
\toprule
Config & Layers & $d_{\text{model}}$ & Heads & Parameters \\
\midrule
tiny   & 4  & 256 & 4  & ${\sim}3.5$M \\
small  & 6  & 384 & 6  & ${\sim}12$M \\
medium & 8  & 512 & 8  & ${\sim}26$M \\
large  & 12 & 768 & 12 & ${\sim}86$M \\
\bottomrule
\end{tabular}
\caption{GPT-2-like model configurations used in the primary experiments (Section~\ref{sec:models}).}
\label{tab:models}
\end{table}

\begin{table}[ht]
\centering\small
\begin{tabular}{lccc>{\raggedright\arraybackslash}p{5cm}}
\toprule
Condition & Correct & Incorrect & Ratio & Purpose \\
\midrule
Equal random   & 1 & 1 random   & 1:1 & Primary: preference against unpredictable errors \\
Equal coherent & 1 & 1 coherent & 1:1 & Control: one consistent false rule, one false string \\
2:1 noise      & 1 & 2 random   & 1:2 & Noise ratio (null $1/3$ on the family share) \\
4:1 noise      & 1 & 4 random   & 1:4 & Noise ratio (null $1/5$ on the family share) \\
\bottomrule
\end{tabular}
\caption{Denoising corpus conditions (Section~\ref{sec:corpus}); each corpus contains 5{,}000 unique problems.}
\label{tab:conditions}
\end{table}

\textbf{Corpus generation example.}\quad Each problem is generated programmatically and verified by SymPy. A representative denoising training example (equal random condition):

\begin{verbatim}
Problem: Expand and simplify 3 * (2x + 5) - 4x
Step 1: 3 * (2x + 5) = 6x + 15
Step 2: 6x + 15 - 4x = 2x + 15
Answer: 2x + 15
\end{verbatim}

The random-error version replaces a derivation step with a plausible but incorrect computation (e.g., \texttt{6x + 15 - 4x = 2x + 11}). In the coherent condition a systematic rule is applied consistently across all problems of the same type (e.g., for distribution, $a(b+c) = ab + c$ instead of $ab + ac$). The paired test set presents the same prompt with both completions for NLL comparison.

\section{The MDL Framing, and What It Predicted}
\label{app:mdl}

This appendix records the reasoning that motivated the experimental design, and what happened to it. All of it describes an idealized two-part MDL learner \citep{rissanen1978modeling,grunwald2007minimum}, not a theorem about gradient-descent training.

\textbf{Setup.}\quad A corpus $D$ of $n$ problems, each appearing with an answer from a correct rule system $T_1$ and one from an alternative $T_2$. The model must encode \emph{both} clusters, which we model as a latent-source mixture with per-problem assignments $z_i \in \{1,2\}$:
\[
L(M) + L(D|M) = \underbrace{L(T_1) + L(T_2)}_{\text{theories}} + \underbrace{L(\{z_i\})}_{\text{source assignments}} + \underbrace{\sum_i L(a_i | T_{z_i})}_{\text{answers given sources}}
\]
A learner that internalizes the more compressible theory encodes its answers cheaply and pays a higher per-token cost for the other cluster.

\textbf{Prediction 1} (random errors): if $T_1$ is compact and $T_2$ is a set of independent random answers, the correct cluster holds more probability mass. \emph{Partly borne out.} That quantity is the aggregate share of Section~\ref{sec:evaluation}. It exceeds $0.5$ at 86M ($59.0\% \pm 0.8\%$) and fails at 3.5M--26M on the denoising corpus ($18.6\%$, $31.6\%$, $44.1\%$), while the standard corpus shows it from 12M (Table~\ref{tab:standard_share}).

\textbf{Prediction 2} (coherent errors): if $L(T_1) \approx L(T_2)$, then at equal frequency the two clusters receive comparable mass. \emph{Borne out} in the coherent conditions we built, at every size, architecture and corpus we tested (Sections~\ref{sec:central}, \ref{sec:learnable}, \ref{sec:wikipedia}, \ref{sec:robustness}, \ref{sec:qwen}); the single-rule $L{=}1$ rung of the ladder, which sits above the null at every size, is the one measured exception (Section~\ref{sec:ladder}).

\textbf{Prediction 3} (multi-rule errors): $N$ compact rules with random assignment to problems cost an extra $O(n \log N)$ bits for the selector, which should restore the correct cluster's advantage. \emph{Not borne out.} Mass on the correct cluster reads $47.9$, $47.8$, $41.8$, $41.9$ and $38.7\%$ at $N = 1, 2, 3, 5, 10$ (Section~\ref{sec:multirule}), and the ladder of Section~\ref{sec:ladder} is flat at 86M ($55.6$, $51.9$, $52.6$, $51.3$).

The graded reading of this framework is not supported by our own measurement, and only its binary consequence (Prediction 2, and Prediction 1 at our largest size) survives. One prediction the graded reading did not make, but the learnability account does, fits the data: the extra cost the mixture pays for the false cluster is the source-assignment term $L(\{z_i\})$, paid only when $z_i$ cannot be recovered from the problem. Making it recoverable, as the learnable selector of Section~\ref{sec:learnable} does, should remove the cost and let the model commit to the false cluster. The family share does drop below the null there while the ladder, which left the assignment unpredictable at every rung, never moved; this is consistent with the reading but does not establish it, since the correct completion still outweighs the selected false string and the below-null share may reflect lower confidence in the correct one (Section~\ref{sec:learnable}). We report compression as an interpretation of the surviving contrast, not as a quantity these experiments measure.

\section{Control Grids in Full}
\label{app:controls}

\subsection{Description-Length Ladder, All Sizes}

\begin{table}[ht]
\centering
\small
\begin{tabular}{llcccc}
\toprule
Params & Support (share, \%, null $50$) & $L{=}1$ & $L{=}4$ & $L{=}16$ & $L{=}64$ \\
\midrule
\multirow{2}{*}{3.5M} & realized & $62.6 \pm 2.2$ & $69.2 \pm 2.0$ & $71.8 \pm 2.0$ & $71.9 \pm 1.2$ \\
 & family & $62.6 \pm 2.2$ & $40.5 \pm 2.5$ & $30.8 \pm 1.4$ & $22.3 \pm 1.2$ \\
\midrule
\multirow{2}{*}{12M} & realized & $63.0 \pm 1.4$ & $72.6 \pm 1.4$ & $80.7 \pm 0.9$ & $81.8 \pm 1.2$ \\
 & family & $63.0 \pm 1.4$ & $49.6 \pm 1.9$ & $45.1 \pm 1.8$ & $36.2 \pm 1.8$ \\
\midrule
\multirow{2}{*}{26M} & realized & $57.5 \pm 1.2$ & $68.0 \pm 0.9$ & $77.4 \pm 1.6$ & $81.6 \pm 1.4$ \\
 & family & $57.5 \pm 1.2$ & $50.8 \pm 1.6$ & $51.8 \pm 1.8$ & $43.9 \pm 1.9$ \\
\midrule
\multirow{2}{*}{86M} & realized & $55.6 \pm 0.7$ & $60.7 \pm 1.8$ & $67.2 \pm 0.9$ & $71.8 \pm 1.2$ \\
 & family & $55.6 \pm 0.7$ & $51.9 \pm 1.6$ & $52.6 \pm 0.6$ & $51.3 \pm 1.3$ \\
\bottomrule
\end{tabular}
\caption{Description-length ladder, 4 rungs $\times$ 4 sizes $\times$ 4 seeds (\%). Completion lengths differ by at most $0.044$ tokens across all 64 runs, and the length-matched subset reproduces these values to within $1.1$ points.}
\label{tab:ladder_full}
\end{table}

\begin{table}[ht]
\centering
\small
\begin{tabular}{lcccc}
\toprule
Support summed (share, \%; ideal $100$) & 3.5M & 12M & 26M & 86M \\
\midrule
Full family, 18 strings & $64.1 \pm 4.4$ & $93.2 \pm 0.6$ & $97.3 \pm 1.3$ & $99.7 \pm 0.0$ \\
Realized, 1 string & $95.1 \pm 1.0$ & $99.6 \pm 0.1$ & $99.9 \pm 0.0$ & $100.0 \pm 0.0$ \\
\bottomrule
\end{tabular}
\caption{Calibration control: the same measures applied to models trained on a corpus containing the same problems and \emph{no false rule anywhere} (2 seeds, \%). The true value is 100 by construction, so the shortfall is the measure's downward bias: $0.3$ points at 86M and $36$ points at 3.5M. The $\pm 4.4$ at 3.5M is a spread of $59.7$ to $68.5$ across two seeds.}
\label{tab:clean_control}
\end{table}

\subsection{Multi-Rule Sweep}

\begin{figure}[ht]
\centering
\includegraphics[width=0.7\textwidth]{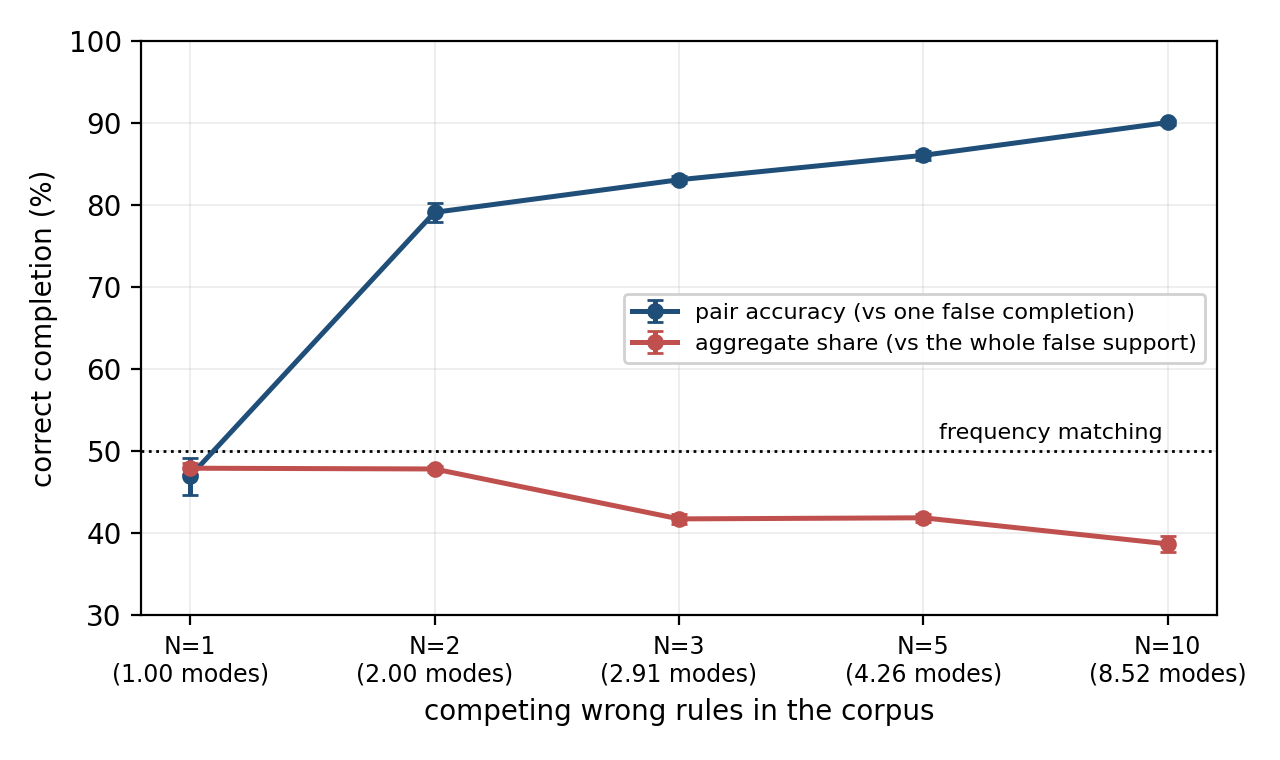}
\caption{Multi-rule errors under both measures (tiny, 4 seeds). Pair accuracy rises monotonically with the number of distinct false modes while the model's allocation between the correct and false clusters does not move at $N{=}1 \to 2$ and declines thereafter. The x-axis is categorical.}
\label{fig:multirule}
\end{figure}

Figure~\ref{fig:multirule} plots the pair-accuracy and aggregate-share columns of Table~\ref{tab:multirule} (Section~\ref{sec:multirule}) against the number of rules.

\subsection{Standard Corpus, Aggregate Share}

\begin{table}[ht]
\centering
\small
\begin{tabular}{lcccc}
\toprule
Aggregate share, \% (null $50$) & 3.5M & 12M & 26M & 86M \\
\midrule
Random (5.51 modes) & $50.2 \pm 0.7$ & $57.6 \pm 0.2$ & $59.0 \pm 0.3$ & $\mathbf{59.4 \pm 0.6}$ \\
\quad equal length & $53.2 \pm 1.1$ & $57.9 \pm 0.4$ & $57.8 \pm 0.2$ & $58.5 \pm 0.7$ \\
Coherent (1 mode) & $47.3 \pm 0.7$ & $49.2 \pm 0.4$ & $50.0 \pm 0.4$ & $50.1 \pm 0.1$ \\
\quad equal length & $49.3 \pm 0.7$ & $49.5 \pm 0.4$ & $49.9 \pm 0.4$ & $49.7 \pm 0.2$ \\
\bottomrule
\end{tabular}
\caption{Aggregate share on the standard (non-denoising) corpus, 4 seeds (\%); $n{=}1101$ problems for random and $915$ for coherent, of which $359$ and $561$ are length-matched. The $59.4$ at 86M is the independent replication of the $59.0$ obtained on the denoising corpus. Per-seed bootstrap intervals for the random condition lie entirely above $0.5$ at 12M, 26M and 86M. The coherent intervals contain $0.5$ from 26M upward; at 3.5M all four lie entirely below it, and at 12M one of four does, consistent with the length account of Section~\ref{sec:central}.}
\label{tab:standard_share}
\end{table}

\section{The Held-Out Scale Grid and the Learnable-Selector Control}
\label{app:scale}

This grid trains one model per condition on 500k texts for 5000 steps at batch 32 ($0.44$ epochs; training windows are sampled with replacement, so some text recurs) and evaluates on ${\approx}2.5$k held-out problems disjoint from training by a hash of the problem parameters. It is the measurement behind Table~\ref{tab:learnable}. Because each text is correct with probability $0.5$ and otherwise carries one false string, the frequency-matching null for the family share is exactly $0.500$ (a marginal correction from arithmetic problems that render identically in every mode raises it to at most $0.513$; we read the band $[0.500, 0.513]$, i.e.\ $50.0$--$51.3\%$). We report the length-matched subset, on which the completion-length confound of Section~\ref{sec:central} is removed.

\textbf{A null of indifference is also a null of competence.}\quad Every null here describes a model that both lacks a preference \emph{and} can compute the correct answer; a model that prefers truth but has not learned the arithmetic reads below it too. So a below-null cell is not on its own evidence of a preference for the false side. Only excursions above the band are unambiguous. This applies to the learnable control too: its random and learnable corpora share a generator and a problem distribution but not their problems, so competence is not matched by construction, and its below-null cells are read as a measured share, not as a demonstrated preference for the false side.

\begin{table}[ht]
\centering\scriptsize
\setlength{\tabcolsep}{2pt}
\begin{tabular}{lcccccccc}
\toprule
Family share, \% (null $50$) & 3.5M & 12M & 26M & 86M & 202M & 413M (GPT-2) & 418M & 474M \\
\midrule
random (eqlen)   & $15.1 \pm 1.7$ & $32.8 \pm 1.1$ & $40.4 \pm 0.6$ & $45.5 \pm 1.4$ & $48.5 \pm 1.0$ & $51.2 \pm 1.1$ & $\mathbf{56.8 \pm 1.1}$ & $\mathbf{54.9 \pm 0.2}$ \\
coherent (eqlen) & $49.4 \pm 0.6$ & $51.6 \pm 0.7$ & $51.2 \pm 0.6$ & $52.4 \pm 0.4$ & $51.9 \pm 0.5$ & $50.5 \pm 0.5$ & $50.2 \pm 0.4$ & $49.9 \pm 0.5$ \\
clean (full)     & --- & --- & --- & $80.9 \pm 0.8$ & $91.7 \pm 0.4$ & $93.1 \pm 0.9$ & $99.4 \pm 0.1$ & $99.2 \pm 0.0$ \\
clean (eqlen)    & --- & --- & --- & $91.0 \pm 0.3$ & $92.2 \pm 0.3$ & $93.8 \pm 0.3$ & $97.9 \pm 0.3$ & $97.2 \pm 0.1$ \\
\bottomrule
\end{tabular}
\caption{Held-out scale grid, family share, \% mass on the correct completion (null $50$); \emph{eqlen} = length-matched subset, \emph{full} = all problems. GPT-2 sizes 3.5M--202M and Qwen3 sizes 418M/474M are 4 seeds on random and coherent; the 413M GPT-2 column and every clean row are 2 seeds. Above-null random cells appear only at the Qwen sizes; coherent stays within about two points of the null throughout. \emph{clean} trains on the same generator with no false rule: its shortfall below $100$ is the measure's downward leak, which cannot lift a value above the ideal.}
\label{tab:scale_grid}
\end{table}

The leak is architecture-dependent: a 413M GPT-2 leaks like a 202M GPT-2 (clean eqlen $93.8$ against $97.9$ for Qwen3 at 418M), and its random cell straddles the null ($52.3$ / $50.1$ by seed), so whether the effect survives in GPT-2 at this size is inconclusive.

\textbf{The learnable-selector control, realized shares.}\quad The false answer is $a{\cdot}b + T[\,s(p)\bmod 10\,]$ with $T$ a fixed 10-entry table, so in the \texttt{learnable}, \texttt{k2} and \texttt{k3} rows the corpus writes its false string from at most ten family members per problem and only the selector $s(p)$ changes among them; \texttt{learnable\_mod} indexes the full random family instead. Every selector row is scored against the same random family (${\approx}17.5$ strings). Family shares for every condition and size are in Table~\ref{tab:learnable}; Table~\ref{tab:learnable_full} gives the \emph{realized} share, the correct completion scored against the single false string the corpus used, which reads how far the model has learned the selector.

\begin{table}[ht]
\centering\small
\begin{tabular}{lccccc}
\toprule
Realized share, \% & 418M & 474M & 1.12B & 2.16B & 3.08B \\
\midrule
\texttt{learnable} (1 digit)   & $66.3 \pm 1.3$ & $65.1 \pm 1.6$ & $65.2 \pm 0.7$ & $71.8 \pm 0.2$ & $63.8 \pm 1.8$ \\
\texttt{k2} (2 digits)         & $85.9 \pm 0.6$ & $88.6 \pm 0.1$ & --- & --- & --- \\
\texttt{k3} (3 digits)         & $89.7 \pm 1.1$ & $91.6 \pm 0.2$ & --- & --- & --- \\
\texttt{learnable\_mod} (all)  & $92.3 \pm 0.1$ & $92.9 \pm 0.2$ & --- & $89.5$ (1 seed) & $93.0 \pm 0.1$ \\
\bottomrule
\end{tabular}
\caption{Realized share (\%, correct completion against the single realized false string) for the learnable-selector conditions. \texttt{learnable} is 4 seeds at 418M and 474M and 2 seeds above; \texttt{k2}, \texttt{k3} and \texttt{learnable\_mod} are 2 seeds except \texttt{learnable\_mod} at 2.16B (1 seed; not run at 1.12B). \texttt{learnable\_mod} indexes the full random family of ${\approx}17.5$ strings rather than the 10-entry table and is never learned. The frequency-matching value for this column is $50$ only for a fitted selector; an unlearned selector over $n$ strings gives $\mathbb{E}[n/(n{+}1)]$ ($90.9$ for the table, ${\approx}94$ for the full family), so the column reads how far the selector was fitted and is not a preference measure. At 1.12B--3.08B the models use bf16 with gradient checkpointing.}
\label{tab:learnable_full}
\end{table}

As the selector depends on more digits it is fitted less and the realized share climbs ($65 \to 89 \to 92$ at 474M), while the family share of Table~\ref{tab:learnable} rises from the below-null value to the null. At the GPT-2 sizes of this grid \texttt{learnable} also sits below the null ($40.9 \pm 0.9$ at 86M, $40.4 \pm 0.9$ at 202M; realized $77.4 \pm 0.0$ and $73.9 \pm 0.2$; 2 seeds), but \texttt{random} is leak-dominated there, so the contrast with \texttt{random} is read at the Qwen sizes and above. From 1.12B upward the full-set clean calibration is $99.5$, $97.8$ and $99.7$ (length-matched $98.1$, $96.4$ and $99.0$ in Table~\ref{tab:learnable}); the single \texttt{learnable\_mod} seed at 2.16B reads $53.2$ and is not averaged with anything.

\textbf{The keyed condition.}\quad The held-out grid also contains a \emph{keyed} condition, in which the single false string per problem is selected by a hash of the problem parameters that the model cannot compute on an unseen problem. Scored against that realized string it reads $92.5 \pm 0.1$ at 418M and $92.7 \pm 0.3$ at 474M (2 seeds), but because the selector is unrecoverable the frequency-matching null for that realized measure is $\mathbb{E}[n/(n{+}1)] \approx 94.1$ over the per-problem family size $n$ (mean $17.5$), not $50$, so these values sit below their null. On the family support, where the null is $50$, keyed is the same measurement as \texttt{random}: the full-set family share is $49.1 \pm 0.8$ at 474M (2 seeds) against $49.6$ for \texttt{random}, and $13.3 \pm 1.3$, $24.6 \pm 1.5$, $30.4 \pm 0.6$ and $37.8 \pm 0.7$ at 3.5M--86M (4 seeds) against $12.7$, $24.1$, $30.0$ and $36.6$. The condition is recorded here for completeness and is not cited anywhere as a positive result.

\section{Compression Measure (gzip)}
\label{app:compression}

To operationalize the MDL argument, we measured compression ratios (gzip level 9) on concatenated correct versus incorrect completions from each paired test set (Table~\ref{tab:compression}).

\begin{table}[ht]
\centering
\small
\begin{tabular}{lcccc}
\toprule
Condition & \multicolumn{3}{c}{gzip ratio} & Pair acc., \% \\
 & correct & incorrect & delta & \\
\midrule
random 50/50     & 0.1627 & 0.1639 & $+0.0012$ & 79.5\% \\
coherent 50/50   & 0.1656 & 0.1658 & $+0.0002$ & 47.2\% \\
contradictory    & 0.1745 & 0.1744 & $-0.0001$ & 49.0\% \\
multirule $N{=}2$  & 0.1671 & 0.1710 & $+0.0038$ & 77.6\% \\
multirule $N{=}3$  & 0.1669 & 0.1722 & $+0.0053$ & 82.8\% \\
multirule $N{=}5$  & 0.1672 & 0.1730 & $+0.0057$ & 84.8\% \\
multirule $N{=}10$ & 0.1676 & 0.1752 & $+0.0076$ & 88.3\% \\
synthetic-world random   & 0.0370 & 0.0516 & $+0.0146$ & 57.7\% \\
synthetic-world coherent & 0.0374 & 0.0384 & $+0.0010$ & 46.6\% \\
\bottomrule
\end{tabular}
\caption{Compression ratio (gzip) and pair accuracy across all 9 conditions (tiny; pair accuracies are 4-seed means, compression ratios are properties of the paired test sets). \emph{contradictory} is a standard-setup condition whose wrong rules are mutually inconsistent (addition overshoots by one, subtraction undershoots by two); the synthetic-world rows are the setting of Appendix~\ref{app:world}. Spearman $\rho = 0.67$; two-sided $p = 0.05$ under the $t$ approximation and $0.06$ by exact permutation, i.e.\ not significant at the conventional level.}
\label{tab:compression}
\end{table}

The multi-rule pair accuracies here ($77.6$, $82.8$, $84.8$, $88.3$) come from the held-out paired test set; Table~\ref{tab:multirule} reports the same checkpoints scored on the enumerated support ($79.1$, $83.1$, $86.1$, $90.1$). Conditions with larger compression gaps show higher pair accuracy, but four of the nine points are the multi-rule conditions, whose pair accuracy is inflated by mass splitting (Section~\ref{sec:multirule}) and whose compression deltas grow with the number of rules because adding rules adds text diversity. The same variable moves both axes, so the correlation cannot be read as evidence that compressibility drives preference; recomputed against aggregate share for the multi-rule points ($47.9$, $47.8$, $41.8$, $41.9$, $38.7$), the positive association does not survive, and its sign reverses (Spearman $\rho = -0.90$ on those five points; exact two-sided $p = 0.08$, so the reversal itself is not established). We retain the appendix as a record of the measurement.

\section{Standard (Non-Denoising) Experiments}
\label{app:standard}

\subsection{Standard Corpus Design}

In the standard setup, each problem appears once with either a correct or incorrect solution. The two groups do not share prompts. ``Matched random'' errors (Section~\ref{sec:robustness}) corrupt the same number of steps as the coherent rule, at random positions with random values; their pair accuracy is $74.5\% \pm 0.4\%$ (tiny, 4 seeds; $74.1$--$75.1$ by seed) against $47.2\%$ for coherent.

\subsection{Standard vs Denoising}

\begin{table}[ht]
\centering
\small
\begin{tabular}{lccc}
\toprule
Pair accuracy, \% & Standard & Denoising & Gap \\
\midrule
tiny   & 79.5\% & 65.3\% & $-14.2$ pp \\
small  & 86.2\% & 74.6\% & $-11.6$ pp \\
medium & 87.1\% & 81.1\% & $-6.0$ pp \\
large  & 88.3\% & 85.2\% & $-3.1$ pp \\
\bottomrule
\end{tabular}
\caption{Standard vs denoising pair accuracy (50/50 random, held-out paired test set, original runs, 4 seeds except the 86M denoising value, 2 seeds; pair accuracy only). The gap closes with scale. The 86M denoising value is the two-seed original run, superseded by the four-seed grid of Table~\ref{tab:central} ($87.4$).}
\label{tab:standard_vs_denoising}
\end{table}

The denoising setup is harder (within-problem contradiction rather than across-corpus mixing).

\subsection{Frequency Effects}

\begin{table}[ht]
\centering
\small
\begin{tabular}{cccc}
\toprule
Proportion & Random, \% & Coherent, \% & False pref., \% \\
\midrule
50/50 & 79.5\% & 47.2\% & 52.8\% \\
40/60 & --- & 27.8\% & 72.2\% \\
30/70 & --- & 14.7\% & 85.3\% \\
20/80 & 63.8\% (1 seed) & 9.6\%  & 90.4\% \\
10/90 & 59.9\% & ---    & --- \\
\bottomrule
\end{tabular}
\caption{Random vs coherent pair accuracy across proportions (standard, tiny; pair accuracy only; 4 seeds unless marked). Coherent: the model follows frequency, and with one false string these values are unaffected by splitting. Random: the column stays above $50\%$ down to 10/90, but on a multi-mode support that is not evidence of a preference (Section~\ref{sec:robustness}). ``False pref.'' $= 1 - \text{coherent pair accuracy}$.}
\label{tab:frequency}
\end{table}

The random 10/90 ($59.9 \pm 0.9$) and 20/80 cells are the re-evaluation of 2026-03-16 recorded in the released per-run files; for 20/80, seed 43 has no evaluation file and seeds 44--45 were not run. The 40/60 and 30/70 random runs exist only in the run registry without retained evaluation files and are not printed.

\subsection{Noise Tolerance}

\begin{table}[ht]
\centering
\small
\begin{tabular}{lccccc}
\toprule
Pair accuracy, \% & Ratio & Tiny & Small & Medium & Large \\
\midrule
Equal random & 1:1 & 65.3\% & 74.6\% & 81.1\% & 85.2\% \\
2:1 noise    & 1:2 & 59.0\% & 68.6\% & 73.6\% & 75.2\% \\
4:1 noise    & 1:4 & 56.6\% & 64.9\% & 66.6\% & 65.8\% \\
\bottomrule
\end{tabular}
\caption{Pair accuracy at increasing noise ratios (denoising, original runs; 4 seeds for 1:1 at 3.5M--26M, 2 seeds for every other cell; pair accuracy only). The number of false completions per problem changes with the ratio, and pair accuracy is sensitive to that, so the decline is descriptive. Table~\ref{tab:kratio} gives the controlled measurement.}
\label{tab:noise}
\end{table}

\begin{table}[ht]
\centering
\small
\begin{tabular}{lccc}
\toprule
86M, 2 seeds & $K{=}1$ (1:1) & $K{=}2$ (1:2) & $K{=}4$ (1:4) \\
\midrule
Frequency-matching null $1/(K{+}1)$ & $50.0$ & $33.3$ & $20.0$ \\
Family share, full set              & $56.3 \pm 0.7$ & $44.9 \pm 0.2$ & $36.1 \pm 0.7$ \\
\quad excess above null             & $+6.3$ & $+11.6$ & $+16.1$ \\
Family share, length-matched ($n{=}1179$) & $57.3 \pm 0.2$ & $40.9 \pm 1.2$ & $31.4 \pm 2.2$ \\
\quad excess above null             & $+7.3$ & $+7.6$ & $+11.4$ \\
Pair accuracy on this support       & $88.7 \pm 0.4$ & $79.4 \pm 0.3$ & $76.3 \pm 0.1$ \\
\bottomrule
\end{tabular}
\caption{Noise ratio on the controlled measure (denoising corpus with $K$ random errors per problem, 86M, 2 seeds, \%). The support is non-strict (4565 problems, mean $8.2$ false modes, length delta $-0.52$ tokens), because a strict filter rejects every problem at $K \geq 2$; the null $1/(K{+}1)$ is exact since each problem appears once correct and $K$ times incorrect. The $K{=}1$ cell is lower than the strict-grid $59.0$ of Table~\ref{tab:central} because the non-strict support keeps problems whose strings collide across problems. The absolute share falls with $K$ as pair accuracy does, but the excess above the null does not shrink.}
\label{tab:kratio}
\end{table}

\subsection{NLL Distribution}

Random errors produce a right-skewed per-pair NLL difference (mean $+0.048$, median $+0.025$, 81.5\% positive; one run per condition, seed 42, standard paired test set). Coherent errors produce a symmetric distribution (45.5\% positive). This is why pair accuracy and the mean per-pair NLL difference can diverge. The coherent distribution's center sits slightly below $0.5$ because the coherent false completions are shorter (by $0.71$ tokens on this standard paired test set, $0.58$ on the denoising corpus); on length-matched problems the aggregate share reads $49.3$--$49.9\%$ on the standard corpus (Table~\ref{tab:standard_share}) and $46.7$--$49.3\%$ on the denoising grid (Section~\ref{sec:central}).

\section{Additional Transfer Experiments}
\label{app:transfer}

\subsection{Wikipedia Entity Substitution, Pair Accuracy}

\begin{table}[ht]
\centering
\small
\begin{tabular}{lcc}
\toprule
\multirow{2}{*}{Size} & Random subst. & Coherent subst. \\
 & pair acc. (uninterpretable) & pair acc. (null $50\%$) \\
\midrule
tiny (3.5M)   & $69.5\% \pm 0.1\%$ & $48.7\% \pm 0.4\%$ \\
small (12M)   & $70.7\% \pm 0.3\%$ & $46.6\% \pm 0.4\%$ \\
medium (26M)  & $71.5\% \pm 0.7\%$ & $46.4\% \pm 0.7\%$ \\
large (86M)   & $71.4\% \pm 0.6\%$ & $45.9\% \pm 1.3\%$ \\
\bottomrule
\end{tabular}
\caption{Wikipedia entity substitution, pair accuracy (50/50, 4 seeds per size, 2{,}000 held-out paired paragraphs). The coherent column has a single false paragraph per prompt and a $50\%$ frequency-matching value; the random column has an unenumerable false support and a frequency-matching value near $100\%$, and is reported without interpretation (Section~\ref{sec:wikipedia}).}
\label{tab:wikipedia}
\end{table}

\subsection{Wikipedia Single-Entity Aggregate Share}

Single-entity paragraphs, absent from training by construction (the corpora kept only paragraphs with at least two entities), have an enumerable false support: the substituted entity type's pool, enumerated in full for the three smallest pools (NORP, LOC, EVENT; a mean of 2764 false strings per problem). Completions are the entity plus at most 60 characters of the shared continuation, which keeps the family-sum null at exactly $0.5$. We evaluate the trained Wikipedia models (PyTorch re-evaluations of the MLX checkpoints) and clean-Wikipedia calibration models trained on the same 20{,}000 paragraphs with no substitution (Table~\ref{tab:wiki_single}).

\begin{table}[ht]
\centering\footnotesize
\setlength{\tabcolsep}{4pt}
\begin{tabular}{lcccc}
\toprule
 & \multicolumn{2}{c}{Single-random (2764 modes, $n{=}468$), full set} & \multicolumn{2}{c}{Single-coherent (1 mode, eqlen $n{=}25$)} \\
\cmidrule(lr){2-3}\cmidrule(lr){4-5}
Share, \% & clean model & random-trained & clean model & coherent-trained \\
\midrule
tiny (3.5M)   & $1.7 \pm 0.1$ & $1.6 \pm 0.1$ & $71.9 \pm 2.9$ & $58.0 \pm 2.2$ \\
small (12M)   & $2.3 \pm 0.2$ & $2.8 \pm 0.2$ & $71.4 \pm 1.2$ & $49.5 \pm 0.3$ \\
medium (26M)  & $4.6 \pm 0.1$ & $3.5 \pm 0.1$ & $71.3 \pm 1.3$ & $49.4 \pm 3.2$ \\
large (86M)   & $5.6 \pm 0.4$ & $4.0 \pm 0.2$ & $68.9 \pm 4.4$ & $52.8 \pm 1.1$ \\
\bottomrule
\end{tabular}
\caption{Wikipedia single-entity aggregate share (\% mass on the correct completion, null $50$, 2 seeds per cell). On the random support a clean model, which should read ${\sim}100$, reads $1.7$--$5.6$: the 2764-mode family sum swamps the single correct string, so the random-trained cells carry no information about preference. On the one-mode coherent support ($n{=}384$ problems) only the length-matched subset (\emph{eqlen}) is read; pooled over the eight coherent-trained runs it is $52.4 \pm 4.0$, at the null, consistent with Table~\ref{tab:wikipedia} but underpowered per cell.}
\label{tab:wiki_single}
\end{table}

\subsection{Synthetic World}
\label{app:world}

In a synthetic world with 50 entities, 4 types and 15 deterministic rules (tiny; pair accuracy only), random 50/50 corruption gives $57.7\% \pm 1.7\%$ and coherent 50/50 corruption $46.6\% \pm 1.8\%$ (4 seeds each). By type, minerals score highest ($68.7\%$) and potions sit near chance ($49.1\%$).

\subsection{Wikipedia Per-Entity-Type}

Pair accuracy is highest for LOC, NORP and GPE and lowest for EVENT and CARDINAL (Table~\ref{tab:wiki_entity}). The per-type samples are small (LOC has 45 pairs, so its $79.4\%$ carries a 95\% binomial interval of roughly $\pm 12$~pp; EVENT has 8), and these are pair-accuracy values on a multi-mode false support, so we do not interpret the ordering.

\begin{table}[ht]
\centering
\small
\begin{tabular}{lcc}
\toprule
Entity type & Pairs & Pair accuracy, \% \\
\midrule
LOC      & 45  & 79.4\% \\
NORP     & 197 & 78.6\% \\
GPE      & 301 & 77.9\% \\
PERSON   & 402 & 72.2\% \\
DATE     & 211 & 68.4\% \\
ORG      & 664 & 66.8\% \\
CARDINAL & 172 & 62.6\% \\
EVENT    & 8   & 56.3\% \\
\bottomrule
\end{tabular}
\caption{Wikipedia pair accuracy by entity type (random substitution; mean over the 16 runs of Table~\ref{tab:wikipedia}, 4 sizes $\times$ 4 seeds; pair accuracy only).}
\label{tab:wiki_entity}
\end{table}

\subsection{Cross-Domain Admixture}

Adding correct cross-domain tasks to a coherent corpus selectively increases derivative pair accuracy ($35\% \to 56\%$ at a 25\% admixture; tiny, pair accuracy only). The effect is non-monotonic (it drops at 50\%, consistent with dilution).

\section{Robustness Checks}
\label{app:robustness}

\subsection{BPE Tokenization}

\begin{table}[ht]
\centering
\small
\begin{tabular}{llcc}
\toprule
Setup & Tokenizer & Random (pair acc., \%) & Coherent (pair acc., \%; null $50$) \\
\midrule
Standard  & Char & 79.5\% & 47.2\% \\
Standard  & BPE  & 85.6\% & 45.9\% \\
Denoising & Char & 65.3\% & 43.5\% \\
Denoising & BPE  & 75.9\% & 49.3\% \\
\bottomrule
\end{tabular}
\caption{BPE vs character-level tokenization (tiny, 4 seeds; pair accuracy only). The coherent column stays at chance under both tokenizers, and the denoising coherent value moves toward $50$ under BPE, consistent with the length account of Section~\ref{sec:central}. The random column rises under BPE, but on a multi-mode support a rise in pair accuracy is not evidence that the preference strengthens; the aggregate share was not measured under BPE.}
\label{tab:bpe}
\end{table}

\subsection{Chained Verification}
\label{app:chained}

Embedding cross-domain verification dependencies in coherent-error tasks raises pair accuracy above the coherent baseline across 10 models (4 tiny, 4 small, 2 large; Table~\ref{tab:chained}).

\begin{table}[ht]
\centering
\small
\begin{tabular}{lccc}
\toprule
Size & Pair accuracy, \% (null $50$) & Seeds & Standard coherent baseline \\
\midrule
tiny (3.5M)  & $70.9\% \pm 1.1\%$ & 4 & $47.2\%$ \\
small (12M)  & $64.2\% \pm 1.4\%$ & 4 & $49.6\%$ \\
large (86M)  & $60.5\% \pm 1.1\%$ & 2 & $51.4\%$ \\
\bottomrule
\end{tabular}
\caption{Chained-verification pair accuracy against the standard-setup coherent baseline measured on the same corpus family (the earlier version of this table printed the \emph{denoising} coherent values here by mistake). Pair accuracy only; the aggregate share was not measured for this condition.}
\label{tab:chained}
\end{table}

The decline with scale should not be read as an inverse-scaling result. Wherever we computed both, the gap between pair accuracy and the aggregate share shrinks with model size ($53.1$, $52.5$, $45.1$ and $28.4$~pp at 3.5M, 12M, 26M and 86M on the denoising corpus), which alone would produce a declining pair-accuracy trend on a condition where only pair accuracy is available. We report the numbers and do not interpret the trend.

\subsection{Generation by Task Type}
\label{app:gentask}

For random-trained 86M models the generative accuracy of Table~\ref{tab:generative} splits by task as differentiation 79.5\%, factoring 70.4\%, equation solving 61.0\% and multi-step arithmetic 0\%; the last requires exact carry propagation across many digits, which autoregressive generation at this scale does not achieve even where paired evaluation discriminates well. Coherent-trained models are lower on every type that is generated at all, with the largest gap on factoring (70.4\% vs 38.3\%) and the smallest on equation solving (61.0\% vs 45.7\%); differentiation reads 79.5\% vs 56.5\%.

\subsection{Learning Curves}

All sizes reach a behavioral plateau by step 3000--4000. In the original large random-trained run (seed 42), pair accuracy on a 500-pair subsample of the standard paired test set reads 90.4\% at step 3000, 86.4\% at 4000 and 87.8\% at 5000. This check rules out an early-stopping artifact; the underlying logs and plotting script are in the released repository.

\section{Per-Seed Details}
\label{app:seeds}

Tables~\ref{tab:seeds_random}--\ref{tab:seeds_coherent} report per-seed pair accuracy for the core denoising conditions as originally run, where the 86M row has two seeds. The control grid of Section~\ref{sec:central} reruns all four sizes with four seeds each on regenerated corpora that match the originals byte for byte (SHA-256 published with the release), and its 86M row therefore supersedes these two-seed entries. Per-run outputs for every cell of both control grids are released; per-seed breakdowns for multi-rule, Wikipedia, and Qwen3 experiments are in the released repository (evaluation logs per model).

\begin{table}[ht]
\centering
\small
\begin{tabular}{lccccc}
\toprule
Size & Seed 1 & Seed 2 & Seed 3 & Seed 4 & Mean \\
\midrule
tiny   & 64.2\% & 64.1\% & 67.0\% & 65.8\% & 65.3\% \\
small  & 75.1\% & 76.3\% & 72.6\% & 74.2\% & 74.6\% \\
medium & 79.6\% & 80.9\% & 82.4\% & 81.3\% & 81.1\% \\
large  & 83.5\% & 86.8\% & ---    & ---    & 85.2\% \\
\bottomrule
\end{tabular}
\caption{Denoising equal random, per-seed pair accuracy (original runs).}
\label{tab:seeds_random}
\end{table}

\begin{table}[ht]
\centering
\small
\begin{tabular}{lccccc}
\toprule
Size & Seed 1 & Seed 2 & Seed 3 & Seed 4 & Mean \\
\midrule
tiny   & 40.2\% & 44.3\% & 46.3\% & 43.0\% & 43.5\% \\
small  & 44.8\% & 45.3\% & 40.3\% & 47.4\% & 44.5\% \\
medium & 45.8\% & 40.9\% & 47.7\% & 48.6\% & 45.8\% \\
large  & 51.6\% & 50.4\% & ---    & ---    & 51.0\% \\
\bottomrule
\end{tabular}
\caption{Denoising equal coherent, per-seed pair accuracy (original runs).}
\label{tab:seeds_coherent}
\end{table}

\clearpage
\section*{NeurIPS Paper Checklist}

\begin{enumerate}

\item {\bf Claims}
    \item[] Question: Do the main claims made in the abstract and introduction accurately reflect the paper's contributions and scope?
    \item[] Answer: \answerYes{}
    \item[] Justification: The primary metric is aggregate share against the enumerated false support, whose null is exactly 0.500 on our corpora. The abstract and introduction state that the effect is established at 86M on two corpora and not below, that pair-accuracy results elsewhere in the paper are uncontrolled and overstate the excess above chance by a factor of three to four, and that the multi-rule and description-length experiments are reported as negative controls rather than as evidence for a mechanism. Claims that appeared in the first version of this work and did not survive their own controls are named and withdrawn in Sections~\ref{sec:multirule}, \ref{sec:ladder} and \ref{sec:discussion}; the learnable-selector control of Section~\ref{sec:learnable} is reported as a measured shift of the family share, with its causal limits stated (unmatched corpora, unresolved source of the below-null share) and compression left as an interpretation.
    \item[] Guidelines:
    \begin{itemize}
        \item The answer \answerNA{} means that the abstract and introduction do not include the claims made in the paper.
        \item The abstract and/or introduction should clearly state the claims made, including the contributions made in the paper and important assumptions and limitations. A \answerNo{} or \answerNA{} answer to this question will not be perceived well by the reviewers.
        \item The claims made should match theoretical and experimental results, and reflect how much the results can be expected to generalize to other settings.
        \item It is fine to include aspirational goals as motivation as long as it is clear that these goals are not attained by the paper.
    \end{itemize}

\item {\bf Limitations}
    \item[] Question: Does the paper discuss the limitations of the work performed by the authors?
    \item[] Answer: \answerYes{}
    \item[] Justification: A dedicated Limitations section discusses the measure's downward leak and architecture dependence, model scale in both directions (1.12B--3.08B replicated at 0.44 epochs; below 86M only generation separates the conditions), domain specificity including the negative single-entity Wikipedia result, the discriminative versus generative gap, and the remaining limits of the causal identification.
    \item[] Guidelines:
    \begin{itemize}
        \item The answer \answerNA{} means that the paper has no limitation while the answer \answerNo{} means that the paper has limitations, but those are not discussed in the paper.
        \item The authors are encouraged to create a separate ``Limitations'' section in their paper.
        \item The paper should point out any strong assumptions and how robust the results are to violations of these assumptions (e.g., independence assumptions, noiseless settings, model well-specification, asymptotic approximations only holding locally). The authors should reflect on how these assumptions might be violated in practice and what the implications would be.
        \item The authors should reflect on the scope of the claims made, e.g., if the approach was only tested on a few datasets or with a few runs. In general, empirical results often depend on implicit assumptions, which should be articulated.
        \item The authors should reflect on the factors that influence the performance of the approach. For example, a facial recognition algorithm may perform poorly when image resolution is low or images are taken in low lighting. Or a speech-to-text system might not be used reliably to provide closed captions for online lectures because it fails to handle technical jargon.
        \item The authors should discuss the computational efficiency of the proposed algorithms and how they scale with dataset size.
        \item If applicable, the authors should discuss possible limitations of their approach to address problems of privacy and fairness.
        \item While the authors might fear that complete honesty about limitations might be used by reviewers as grounds for rejection, a worse outcome might be that reviewers discover limitations that aren't acknowledged in the paper. The authors should use their best judgment and recognize that individual actions in favor of transparency play an important role in developing norms that preserve the integrity of the community. Reviewers will be specifically instructed to not penalize honesty concerning limitations.
    \end{itemize}

\item {\bf Theory assumptions and proofs}
    \item[] Question: For each theoretical result, does the paper provide the full set of assumptions and a complete (and correct) proof?
    \item[] Answer: \answerNA{}
    \item[] Justification: The paper does not claim formal theoretical results. Appendix~\ref{app:mdl} presents three informal MDL-motivated predictions for an idealized learner, without proofs, explicitly marked as motivation rather than as theorems about neural network training. The paper is primarily empirical.
    \item[] Guidelines:
    \begin{itemize}
        \item The answer \answerNA{} means that the paper does not include theoretical results.
        \item All the theorems, formulas, and proofs in the paper should be numbered and cross-referenced.
        \item All assumptions should be clearly stated or referenced in the statement of any theorems.
        \item The proofs can either appear in the main paper or the supplemental material, but if they appear in the supplemental material, the authors are encouraged to provide a short proof sketch to provide intuition.
        \item Inversely, any informal proof provided in the core of the paper should be complemented by formal proofs provided in appendix or supplemental material.
        \item Theorems and Lemmas that the proof relies upon should be properly referenced.
    \end{itemize}

    \item {\bf Experimental result reproducibility}
    \item[] Question: Does the paper fully disclose all the information needed to reproduce the main experimental results of the paper to the extent that it affects the main claims and/or conclusions of the paper (regardless of whether the code and data are provided or not)?
    \item[] Answer: \answerYes{}
    \item[] Justification: All hyperparameters, model architectures, data generation procedures, evaluation protocols, and random seeds are specified in Section~\ref{sec:methodology} and Appendix~\ref{app:reproducibility}; more than 500 training runs are documented. Full generation templates, the support-enumeration scripts and per-run metric JSON are in the public repository.
    \item[] Guidelines:
    \begin{itemize}
        \item The answer \answerNA{} means that the paper does not include experiments.
        \item If the paper includes experiments, a \answerNo{} answer to this question will not be perceived well by the reviewers: Making the paper reproducible is important, regardless of whether the code and data are provided or not.
        \item If the contribution is a dataset and\slash or model, the authors should describe the steps taken to make their results reproducible or verifiable.
        \item Depending on the contribution, reproducibility can be accomplished in various ways. For example, if the contribution is a novel architecture, describing the architecture fully might suffice, or if the contribution is a specific model and empirical evaluation, it may be necessary to either make it possible for others to replicate the model with the same dataset, or provide access to the model. In general. releasing code and data is often one good way to accomplish this, but reproducibility can also be provided via detailed instructions for how to replicate the results, access to a hosted model (e.g., in the case of a large language model), releasing of a model checkpoint, or other means that are appropriate to the research performed.
        \item While NeurIPS does not require releasing code, the conference does require all submissions to provide some reasonable avenue for reproducibility, which may depend on the nature of the contribution. For example
        \begin{enumerate}
            \item If the contribution is primarily a new algorithm, the paper should make it clear how to reproduce that algorithm.
            \item If the contribution is primarily a new model architecture, the paper should describe the architecture clearly and fully.
            \item If the contribution is a new model (e.g., a large language model), then there should either be a way to access this model for reproducing the results or a way to reproduce the model (e.g., with an open-source dataset or instructions for how to construct the dataset).
            \item We recognize that reproducibility may be tricky in some cases, in which case authors are welcome to describe the particular way they provide for reproducibility. In the case of closed-source models, it may be that access to the model is limited in some way (e.g., to registered users), but it should be possible for other researchers to have some path to reproducing or verifying the results.
        \end{enumerate}
    \end{itemize}

\item {\bf Open access to data and code}
    \item[] Question: Does the paper provide open access to the data and code, with sufficient instructions to faithfully reproduce the main experimental results, as described in supplemental material?
    \item[] Answer: \answerYes{}
    \item[] Justification: The public repository \url{https://github.com/Rai220/compression-drives-truth} contains the code for data generation, training and evaluation, the corpus generators and support-enumeration scripts, per-run metric JSON for all cluster grids, and SHA-256 checksums of the regenerable corpora.
    \item[] Guidelines:
    \begin{itemize}
        \item The answer \answerNA{} means that paper does not include experiments requiring code.
        \item Please see the NeurIPS code and data submission guidelines (\url{https://neurips.cc/public/guides/CodeSubmissionPolicy}) for more details.
        \item While we encourage the release of code and data, we understand that this might not be possible, so \answerNo{} is an acceptable answer. Papers cannot be rejected simply for not including code, unless this is central to the contribution (e.g., for a new open-source benchmark).
        \item The instructions should contain the exact command and environment needed to run to reproduce the results. See the NeurIPS code and data submission guidelines (\url{https://neurips.cc/public/guides/CodeSubmissionPolicy}) for more details.
        \item The authors should provide instructions on data access and preparation, including how to access the raw data, preprocessed data, intermediate data, and generated data, etc.
        \item The authors should provide scripts to reproduce all experimental results for the new proposed method and baselines. If only a subset of experiments are reproducible, they should state which ones are omitted from the script and why.
        \item At submission time, to preserve anonymity, the authors should release anonymized versions (if applicable).
        \item Providing as much information as possible in supplemental material (appended to the paper) is recommended, but including URLs to data and code is permitted.
    \end{itemize}

\item {\bf Experimental setting/details}
    \item[] Question: Does the paper specify all the training and test details (e.g., data splits, hyperparameters, how they were chosen, type of optimizer) necessary to understand the results?
    \item[] Answer: \answerYes{}
    \item[] Justification: See Section~\ref{sec:models} (optimizer, lr, schedule, batch size, steps), Section~\ref{sec:qwen} (Qwen3 and FineWeb-Edu run details) and Appendix~\ref{app:reproducibility} (corpus examples, hardware). Train/test separation is by construction: paired test sets are generated with a separate seed (999), the held-out scale grid splits by a hash of the problem parameters, and the denoising aggregate share is measured on training problems where the null is exact.
    \item[] Guidelines:
    \begin{itemize}
        \item The answer \answerNA{} means that the paper does not include experiments.
        \item The experimental setting should be presented in the core of the paper to a level of detail that is necessary to appreciate the results and make sense of them.
        \item The full details can be provided either with the code, in appendix, or as supplemental material.
    \end{itemize}

\item {\bf Experiment statistical significance}
    \item[] Question: Does the paper report error bars suitably and correctly defined or other appropriate information about the statistical significance of the experiments?
    \item[] Answer: \answerYes{}
    \item[] Justification: All $\pm$ values are population standard deviations (ddof $=0$) across random seeds, not standard errors. Pair-accuracy evaluations store a one-sided Wilcoxon signed-rank $p$ on per-pair NLL differences and a bootstrap interval, released with the per-run outputs; per-run 95\% confidence intervals are bootstrap intervals over problems with 5000 resamples (Section~\ref{sec:evaluation}). The number of seeds is reported for every condition. Per-seed breakdowns for the core denoising conditions are provided in Appendix~\ref{app:seeds}. For the 490M FineWeb-Edu experiment (2 seeds), the reported values are pair accuracy on an enumeration-free support and are marked as descriptive.
    \item[] Guidelines:
    \begin{itemize}
        \item The answer \answerNA{} means that the paper does not include experiments.
        \item The authors should answer \answerYes{} if the results are accompanied by error bars, confidence intervals, or statistical significance tests, at least for the experiments that support the main claims of the paper.
        \item The factors of variability that the error bars are capturing should be clearly stated (for example, train/test split, initialization, random drawing of some parameter, or overall run with given experimental conditions).
        \item The method for calculating the error bars should be explained (closed form formula, call to a library function, bootstrap, etc.)
        \item The assumptions made should be given (e.g., Normally distributed errors).
        \item It should be clear whether the error bar is the standard deviation or the standard error of the mean.
        \item It is OK to report 1-sigma error bars, but one should state it. The authors should preferably report a 2-sigma error bar than state that they have a 96\% CI, if the hypothesis of Normality of errors is not verified.
        \item For asymmetric distributions, the authors should be careful not to show in tables or figures symmetric error bars that would yield results that are out of range (e.g., negative error rates).
        \item If error bars are reported in tables or plots, the authors should explain in the text how they were calculated and reference the corresponding figures or tables in the text.
    \end{itemize}

\item {\bf Experiments compute resources}
    \item[] Question: For each experiment, does the paper provide sufficient information on the computer resources (type of compute workers, memory, time of execution) needed to reproduce the experiments?
    \item[] Answer: \answerYes{}
    \item[] Justification: See Appendix~\ref{app:reproducibility}. Original runs: approximately 150 GPU-hours (an estimate carried from the first version; no per-run timing was retained) on Modal.com T4/A10G and Kaggle T4 GPUs plus approximately 100 hours of Apple M4 Pro (MLX) training over 218 local runs. The two control grids and the held-out scale grid ran on 8${\times}$H100 nodes (per-run timing not recorded). The August 2026 follow-up program (billion-scale grid, learnable-selector conditions, clean controls, $K$-ratio sweep, reruns) took approximately 45 single-H100 hours over 81 logged runs. Hardware is specified per experiment type.
    \item[] Guidelines:
    \begin{itemize}
        \item The answer \answerNA{} means that the paper does not include experiments.
        \item The paper should indicate the type of compute workers CPU or GPU, internal cluster, or cloud provider, including relevant memory and storage.
        \item The paper should provide the amount of compute required for each of the individual experimental runs as well as estimate the total compute.
        \item The paper should disclose whether the full research project required more compute than the experiments reported in the paper (e.g., preliminary or failed experiments that didn't make it into the paper).
    \end{itemize}

\item {\bf Code of ethics}
    \item[] Question: Does the research conducted in the paper conform, in every respect, with the NeurIPS Code of Ethics \url{https://neurips.cc/public/EthicsGuidelines}?
    \item[] Answer: \answerYes{}
    \item[] Justification: The research uses only synthetic data, publicly available Wikipedia text and the public FineWeb-Edu corpus. No human subjects, no private data, no dual-use concerns beyond those discussed in Ethical Considerations.
    \item[] Guidelines:
    \begin{itemize}
        \item The answer \answerNA{} means that the authors have not reviewed the NeurIPS Code of Ethics.
        \item If the authors answer \answerNo, they should explain the special circumstances that require a deviation from the Code of Ethics.
        \item The authors should make sure to preserve anonymity (e.g., if there is a special consideration due to laws or regulations in their jurisdiction).
    \end{itemize}

\item {\bf Broader impacts}
    \item[] Question: Does the paper discuss both potential positive societal impacts and negative societal impacts of the work performed?
    \item[] Answer: \answerYes{}
    \item[] Justification: The Ethical Considerations section discusses both the defensive value (informing data curation) and the risk (characterizing conditions under which misinformation evades filtering).
    \item[] Guidelines:
    \begin{itemize}
        \item The answer \answerNA{} means that there is no societal impact of the work performed.
        \item If the authors answer \answerNA{} or \answerNo, they should explain why their work has no societal impact or why the paper does not address societal impact.
        \item Examples of negative societal impacts include potential malicious or unintended uses (e.g., disinformation, generating fake profiles, surveillance), fairness considerations (e.g., deployment of technologies that could make decisions that unfairly impact specific groups), privacy considerations, and security considerations.
        \item The conference expects that many papers will be foundational research and not tied to particular applications, let alone deployments. However, if there is a direct path to any negative applications, the authors should point it out. For example, it is legitimate to point out that an improvement in the quality of generative models could be used to generate Deepfakes for disinformation. On the other hand, it is not needed to point out that a generic algorithm for optimizing neural networks could enable people to train models that generate Deepfakes faster.
        \item The authors should consider possible harms that could arise when the technology is being used as intended and functioning correctly, harms that could arise when the technology is being used as intended but gives incorrect results, and harms following from (intentional or unintentional) misuse of the technology.
        \item If there are negative societal impacts, the authors could also discuss possible mitigation strategies (e.g., gated release of models, providing defenses in addition to attacks, mechanisms for monitoring misuse, mechanisms to monitor how a system learns from feedback over time, improving the efficiency and accessibility of ML).
    \end{itemize}

\item {\bf Safeguards}
    \item[] Question: Does the paper describe safeguards that have been put in place for responsible release of data or models that have a high risk for misuse (e.g., pre-trained language models, image generators, or scraped datasets)?
    \item[] Answer: \answerNA{}
    \item[] Justification: All models are trained from scratch on synthetic math or on filtered web text with synthetic math; the largest is 3.08B parameters and is trained on synthetic math only. They have no capabilities beyond the narrow experimental domain and pose no misuse risk.
    \item[] Guidelines:
    \begin{itemize}
        \item The answer \answerNA{} means that the paper poses no such risks.
        \item Released models that have a high risk for misuse or dual-use should be released with necessary safeguards to allow for controlled use of the model, for example by requiring that users adhere to usage guidelines or restrictions to access the model or implementing safety filters.
        \item Datasets that have been scraped from the Internet could pose safety risks. The authors should describe how they avoided releasing unsafe images.
        \item We recognize that providing effective safeguards is challenging, and many papers do not require this, but we encourage authors to take this into account and make a best faith effort.
    \end{itemize}

\item {\bf Licenses for existing assets}
    \item[] Question: Are the creators or original owners of assets (e.g., code, data, models), used in the paper, properly credited and are the license and terms of use explicitly mentioned and properly respected?
    \item[] Answer: \answerYes{}
    \item[] Justification: Wikipedia data (2023-11-01 dump) is used under CC BY-SA 4.0. FineWeb-Edu \citep{penedo2024fineweb} is used under ODC-By 1.0. All software dependencies (PyTorch, MLX, SymPy, spaCy) are open source.
    \item[] Guidelines:
    \begin{itemize}
        \item The answer \answerNA{} means that the paper does not use existing assets.
        \item The authors should cite the original paper that produced the code package or dataset.
        \item The authors should state which version of the asset is used and, if possible, include a URL.
        \item The name of the license (e.g., CC-BY 4.0) should be included for each asset.
        \item For scraped data from a particular source (e.g., website), the copyright and terms of service of that source should be provided.
        \item If assets are released, the license, copyright information, and terms of use in the package should be provided. For popular datasets, \url{paperswithcode.com/datasets} has curated licenses for some datasets. Their licensing guide can help determine the license of a dataset.
        \item For existing datasets that are re-packaged, both the original license and the license of the derived asset (if it has changed) should be provided.
        \item If this information is not available online, the authors are encouraged to reach out to the asset's creators.
    \end{itemize}

\item {\bf New assets}
    \item[] Question: Are new assets introduced in the paper well documented and is the documentation provided alongside the assets?
    \item[] Answer: \answerYes{}
    \item[] Justification: Code, data generation scripts and per-run outputs are documented in the public repository's README and inline comments.
    \item[] Guidelines:
    \begin{itemize}
        \item The answer \answerNA{} means that the paper does not release new assets.
        \item Researchers should communicate the details of the dataset\slash code\slash model as part of their submissions via structured templates. This includes details about training, license, limitations, etc.
        \item The paper should discuss whether and how consent was obtained from people whose asset is used.
        \item At submission time, remember to anonymize your assets (if applicable). You can either create an anonymized URL or include an anonymized zip file.
    \end{itemize}

\item {\bf Crowdsourcing and research with human subjects}
    \item[] Question: For crowdsourcing experiments and research with human subjects, does the paper include the full text of instructions given to participants and screenshots, if applicable, as well as details about compensation (if any)?
    \item[] Answer: \answerNA{}
    \item[] Justification: No human subjects or crowdsourcing involved. All data is generated programmatically.
    \item[] Guidelines:
    \begin{itemize}
        \item The answer \answerNA{} means that the paper does not involve crowdsourcing nor research with human subjects.
        \item Including this information in the supplemental material is fine, but if the main contribution of the paper involves human subjects, then as much detail as possible should be included in the main paper.
        \item According to the NeurIPS Code of Ethics, workers involved in data collection, curation, or other labor should be paid at least the minimum wage in the country of the data collector.
    \end{itemize}

\item {\bf Institutional review board (IRB) approvals or equivalent for research with human subjects}
    \item[] Question: Does the paper describe potential risks incurred by study participants, whether such risks were disclosed to the subjects, and whether Institutional Review Board (IRB) approvals (or an equivalent approval/review based on the requirements of your country or institution) were obtained?
    \item[] Answer: \answerNA{}
    \item[] Justification: No human subjects involved.
    \item[] Guidelines:
    \begin{itemize}
        \item The answer \answerNA{} means that the paper does not involve crowdsourcing nor research with human subjects.
        \item Depending on the country in which research is conducted, IRB approval (or equivalent) may be required for any human subjects research. If you obtained IRB approval, you should clearly state this in the paper.
        \item We recognize that the procedures for this may vary significantly between institutions and locations, and we expect authors to adhere to the NeurIPS Code of Ethics and the guidelines for their institution.
        \item For initial submissions, do not include any information that would break anonymity (if applicable), such as the institution conducting the review.
    \end{itemize}

\item {\bf Declaration of LLM usage}
    \item[] Question: Does the paper describe the usage of LLMs if it is an important, original, or non-standard component of the core methods in this research? Note that if the LLM is used only for writing, editing, or formatting purposes and does \emph{not} impact the core methodology, scientific rigor, or originality of the research, declaration is not required.
    \item[] Answer: \answerNA{}
    \item[] Justification: LLMs (Claude) were used for writing assistance, code editing, and bibliography verification, but not as a component of the research methodology. All experimental code, data generation, model training, and evaluation were implemented and verified independently of LLM assistance.
    \item[] Guidelines:
    \begin{itemize}
        \item The answer \answerNA{} means that the core method development in this research does not involve LLMs as any important, original, or non-standard components.
        \item Please refer to our LLM policy in the NeurIPS handbook for what should or should not be described.
    \end{itemize}

\end{enumerate}

\end{document}